\documentclass[11pt]{article}
	
\usepackage{graphicx}
\usepackage{bm}
\usepackage{fancyhdr,graphicx,amsmath,amssymb}
\usepackage{subfigure}
\usepackage{xcolor}	
\usepackage[linesnumbered,ruled]{algorithm2e}
\include{pythonlisting}
\usepackage{natbib} 
\usepackage{url} %
\usepackage{mathtools}
\usepackage{float}
\newtheorem{prop}{Proposition}

\newtheorem{theorem}{Theorem}
\DeclareMathOperator*{\argmax}{argmax} 
\usepackage{enumitem}

	\newcommand{\blind}{0}
	
    \makeatletter
    \renewcommand\section{\@startsection {section}{1}{\z@}%
                                       {-3.5ex \@plus -1ex \@minus -.2ex}%
                                       {2.3ex \@plus.2ex}%
                                       {\normalfont\fontfamily{phv}\fontsize{16}{19}\bfseries}}
    \renewcommand\subsection{\@startsection{subsection}{2}{\z@}%
                                         {-3.25ex\@plus -1ex \@minus -.2ex}%
                                         {1.5ex \@plus .2ex}%
                                         {\normalfont\fontfamily{phv}\fontsize{14}{17}\bfseries}}
    \renewcommand\subsubsection{\@startsection{subsubsection}{3}{\z@}%
                                        {-3.25ex\@plus -1ex \@minus -.2ex}%
                                         {1.5ex \@plus .2ex}%
                                         {\normalfont\normalsize\fontfamily{phv}\fontsize{14}{17}\selectfont}}
    \makeatother
	
	\usepackage{amsmath}
	\usepackage{graphicx}
	\usepackage{natbib} 
	\usepackage{url} 
	
\begin{document}
		
		\def\spacingset#1{\renewcommand{\baselinestretch}%
			{#1}\small\normalsize} \spacingset{1}
		
		\if0\blind
		{
			\title{\bf Federated Multilinear Principal Component Analysis with Applications in Prognostics}
			\author{
   Chengyu Zhou\\Edward P. Fitts Department of Industrial and Systems Engineering,\\ North Carolina State University \vspace{3mm} \\ Yuqi Su\\ Operations Research Program, \\North Carolina State University \vspace{3mm}\\ Tangbin Xia\\ Department of Industrial Engineering, \\Shanghai Jiao Tong University \vspace{3mm}\\ Xiaolei Fang*  \\Edward P. Fitts Department of Industrial and Systems Engineering,\\North Carolina State University
		 }
			\date{}
			\maketitle
		} \fi
		
		\if1\blind
		{

            \title{\bf Federated Multilinear Principal Component Analysis with Applications in Prognostics}
			
\bigskip
			\bigskip
			\bigskip
			\begin{center}
				{\LARGE\bf Federated Multilinear Principal Component Analysis with Applications in Prognostics }
			\end{center}
			\medskip
		} \fi
		\bigskip
		
	\begin{abstract}

Multilinear Principal Component Analysis (MPCA) is a widely utilized method for the dimension reduction of tensor data. However, the integration of MPCA into federated learning remains unexplored in existing research. To tackle this gap, this article proposes a Federated Multilinear Principal Component Analysis (FMPCA) method, which enables multiple users to collaboratively reduce the dimension of their tensor data while keeping each user's data local and confidential. The proposed FMPCA method is guaranteed to have the same performance as traditional MPCA. An application of the proposed FMPCA in industrial prognostics is also demonstrated. Simulated data and a real-world data set are used to validate the performance of the proposed method.

	\end{abstract}
			
	\noindent%

	\spacingset{1.32} 

\newpage

\section{Introduction} \label{s:intro}

\begin{figure*}[!htp]
\centering
 \includegraphics[width=0.5\textwidth]{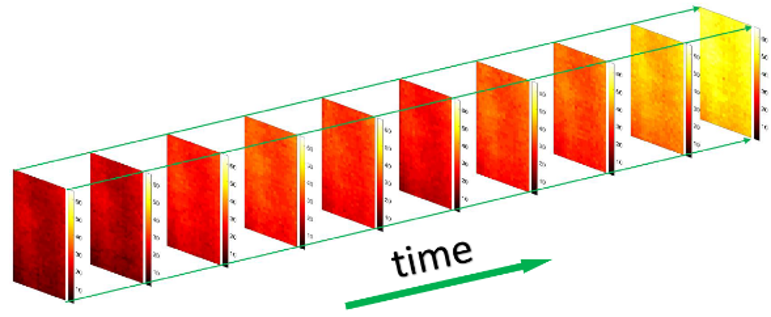}
 \caption{An image stream.}
 \label{fig: ImageSteam}
\end{figure*}

The use of tensors is progressively widespread in the realms of data analytics and machine learning. As an extension of vectors and matrices, a tensor is a multi-dimensional array of numbers that provides a means to represent data across multiple dimensions. As an illustration, Figure \ref{fig: ImageSteam} shows an image stream that can be seen as a three-dimensional tensor, where the first two dimensions denote the pixels within each image, while the third dimension represents the distinct images in the sequence. One of the advantages of representing data as a tensor, as opposed to reshaping it into a vector or matrix, lies in its ability to capture intricate relationships within the data, especially when interactions occur across multiple dimensions. For instance, the image stream depicted in Figure \ref{fig: ImageSteam} exhibits a spatiotemporal correlation structure. Specifically, pixels within each image have spatial correlation, and pixels at the same location across multiple images are temporally correlated. Transforming the image stream into a vector or matrix would disrupt the spatiotemporal correlation structure, whereas representing it as a three-dimensional tensor preserves this correlation. In addition to capturing intricate relationships, other benefits of using tensors include compatibility with multi-modal data (i.e., accommodating diverse types of data in a unified structure) and facilitating parallel processing (i.e., enabling the parallelization of operations), etc. As a result, the volume of research in tensor-based data analytics has been rapidly increasing in recent years \citep{shen2022super,gahrooei2021multiple,yan2019structured,hu2023personalized,zhou2023supervised, zhen2023image,zhang2023tensor}.

A common challenge in tensor-based data analytics is that tensor data is typically ultrahigh dimensional. Taking the image stream in Figure \ref{fig: ImageSteam} as an example again, if there are $1000$ images and the resolution of each image is $1024\times 1024$, the total number of pixels is more than one billion. As a result, dimension reduction techniques are typically employed to decrease the dimension of tensors before the construction of data analytics models. One widely used dimension reduction method for tensor data is Multilinear Principal Component Analysis (MPCA) \citep{lu2008mpca}, which is an extension of traditional Principal Component Analysis (PCA) to handle multi-dimensional data. While PCA is designed for matrices, MPCA generalizes the idea to higher-order tensors. It aims to extract the most important features and reduce the dimensionality of multi-dimensional data while considering interactions between different dimensions. Mathematically, MPCA is expressed as an optimization problem aiming to determine a set of projection matrices. These matrices are designed to efficiently project high-dimensional tensors onto a lower-dimensional tensor subspace, with the objective of maximizing the variation of the projected tensors within this subspace. MPCA has the potential to substantially decrease dimensionality while retaining essential information of tensor data. For instance, it could assist in reducing the dimension of the tensor in Figure \ref{fig: ImageSteam} from $1024\times 1024\times 1000$ to a more manageable size, say $10\times 10\times 20$.  As a result, it has been extensively applied in various data analytics models, including tasks like classification \citep{su2018efficient,wang2010multilinear}, clustering \citep{he2017pattern,he2019unsupervised}, regression \citep{fang2019image,liu2018incremental}, and beyond.

Although there is substantial research utilizing MPCA, there has been no exploration of its integration with Federated Learning (FL). FL has been a hot research topic ever since the concept was proposed by Google in 2016 \citep{konevcny2016federated, yang2019federated,briggs2021review,kontar2021internet}. It comprises a family of algorithms that facilitate model training across numerous decentralized edge devices (such as smartphones, IoT devices, or local servers) holding local data samples. FL plays a vital role in addressing the challenge of \textit{limited data availability} in data analytics since it enables multiple organizations to use their isolated data to jointly train data analytics models while keeping each participant’s data local and confidential. This is exceptionally valuable for applications where a single organization does not have enough data to independently train a reliable model, and the data from multiple organizations cannot be simply shared and merged due to privacy constraints. Since numerous applications dealing with tensor data also encounter challenges related to limited data availability, it is advantageous to investigate how to perform MPCA within the framework of federated learning. However, the integration of MPCA with FL is a nontrivial task. The traditional optimization algorithm for MPCA, as proposed by \cite{lu2008mpca}, is an iterative approach designed to identify projection matrices capable of projecting high-dimensional tensors onto a low-dimensional tensor subspace while retaining critical information from these high-dimensional tensors. It consists of three steps: \textit{Preprocessing}, \textit{Initialization}, and \textit{Local Optimization}, all of which fail to ensure the data privacy of participants. This is because they require each user to compute intermediate results using their local data and share them with the server that coordinates the computation of all users, and the intermediate results expose information from the local data (more details will be provided in Section \ref{sec: mpcaleakage}). 

To address the aforementioned challenge, this article proposes a Federated Multilinear Principal Component Analysis (FMPCA) method that allows multiple users to jointly perform MPCA while keeping each user's data local and confidential. The proposed FMPCA method is built upon three newly devised federated learning algorithms, each tailored for one of the three steps of MPCA (i.e., \textit{Preprocessing}, \textit{Initialization}, and \textit{Local Optimization}). Specifically, we first propose {\fontfamily{qcr}\selectfont Federated Centralization Algorithm} for the \textit{Preprocessing} step of MPCA. It enables all participants to collaboratively compute a global mean of their tensor data without compromising data privacy. This is achieved by letting each user calculate a local mean using its data and mask it with specially designed perturbation. The server aggregates the perturbed local means, and the aggregation cancels out perturbations from different users and yields the global mean. Next, we develop {\fontfamily{qcr}\selectfont Federated Initialization Algorithm} and {\fontfamily{qcr}\selectfont Federated Local Optimization Algorithm} for the \textit{Initialization} and \textit{Local Optimization} steps of MPCA, respectively. The two algorithms enable multiple users to jointly initialize and update the projection matrices of the MPCA algorithm until convergence. Both are incremental algorithms where the first user calculates local projection matrices and shares them with the next user, who updates the projection matrices using its data. This process iteratively continues until all users are involved in updating the projection matrices using their local data. We will prove that all three proposed algorithms achieve identical performance to the three steps in the classic MPCA algorithm as presented in \cite{lu2008mpca}. In other words, the proposed FMPCA method can protect users' data privacy while achieving the same performance as applying MPCA on the data aggregated from all users. 

To validate the performance of the proposed FMPCA method, this article applies it to industrial prognostics.  Industrial prognostics focuses on utilizing the degradation signal of an engineering asset to predict its time-to-failure (TTF). Degradation is an irreversible process of damage accumulation that eventually results in the failure of engineering systems. While the underlying degradation process of engineering systems is typically not directly observable, it is often associated with some manifestations that can be monitored through sensors. The sensing data related to the degradation processes are usually referred to as degradation signals. Industrial prognostics work by establishing a function that maps the TTFs of engineering assets to their degradation signals. In recent years, there has been a growing trend in the utilization of imaging sensing technology for monitoring the condition of engineering systems, which generates image-based degradation signals. This is attributed to the noncontact nature of imaging sensing technology, making it more accessible for implementation. Also, imaging-based degradation signals offer more comprehensive information compared to time series-based signals. In this article, we consider a federated industrial prognostic model for applications with imaging data. Specifically, multiple users will first employ the proposed FMPCA method to jointly perform MPCA while keeping their data local and confidential, which provides low-dimensional features. Next, a prognostic model based on federated (log)-location-scale (LLS) regression \citep{su2022federated} is established by regressing TTF against the features from FMPCA. 

The remainder of this paper is structured as follows. Section \ref{sec:preliminary} introduces basic notations, tensor symbols, and fundamental operators in multilinear algebra. In Section \ref{s:metho}, we provide a review of MPCA as well as its optimization algorithm and discuss challenges related to data privacy protection. Section \ref{sec: FMPCA} details the proposed FMPCA method, and Section \ref{sec:prognostics} explores the application of FMPCA in industrial prognostics. Sections \ref{sec:sim} and \ref{sec:case} validate the effectiveness of the proposed method using a simulated dataset and data from rotating machinery, respectively. Finally, Section \ref{sec:conclusion} presents the conclusion.

\section{Preliminaries}\label{sec:preliminary}
In this section, we will introduce some basic notations and definitions of tensor symbols defined throughout this paper. Vectors are denoted by lowercase boldface letters, e.g., $\bm{x}$, matrices are denoted by boldface uppercase letters, e.g., $\bm{X}$, and tensors are denoted by calligraphic letters, e.g., $\mathcal{X}$. The \textit{order} of a tensor is the number of dimensions, also known as \textit{modes} or \textit{ways}. Indices are denoted by lowercase letters and span the range from 1 to the uppercase letter of the index, e.g., $m = 1, 2, \ldots, M$. An Nth-order tensor is denoted as $\mathcal{X} \in \mathbb{R}^{I_1 \times I_2 \times    \times I_N}$, where $I_n$ represents the $n$th mode of $\mathcal{X}$. The $(i_1, i_2, \ldots , i_N)$th entry of $ \mathcal{X} \in \mathbb{R}^{I_1 \times I_2 \times    \times I_N} $ is denoted by $x_{i_1,i_2,\ldots,i_n}$. A fiber of $\mathcal{X}$ is a vector defined by fixing every index but one. A matrix column is a mode-1 fiber and a matrix row is a mode-2 fiber. The \textit{vectorization} of $\mathcal{X}$, denoted by \textit{vec}($\mathcal{X}$), is obtained by stacking all mode-1 fibers of $\mathcal{X}$. The mode-n \textit{matricization} of a tensor $\mathcal{X} \in \mathbb{R}^{I_1 \times I_2 \times    \times I_N }$, is a matrix whose columns are mode-n \textit{matricization} of $\mathcal{X}$ in the lexicographical order. The mode-n product of a tensor $\mathcal{X} \in \mathbb{R}^{I_1 \times I_2 \times    \times I_N}$ and a matrix $\bm{U}_{n} \in \mathbb{R}^{J_n \times I_n}$, denoted by $\mathcal{X} \times_{n} \bm{U}_{n}$, is a tensor whose entry is $(\mathcal{X} \times_{n} \bm{U}_{n})_{i_{1}, \ldots, i_{n-1}, j_{n}, i_{n+1}, \ldots, i_{N}} = \sum_{I_n = 1}^{I_N } x_{i_1,\ldots,i_N} u_{j, i_n}$. The inner product of two tensors $\mathcal{A}, \mathcal{B} \in \mathbb{R}^{I_1 \times I_2 \times \ldots \times I_N}$ is denoted by $\langle \mathcal{A}, \mathcal{B} \rangle = \sum_{i_1, \dots, i_N} a_{i_1, \dots, i_N} b_{i_1, \ldots, i_N}$ and the Frobenius norm of $\mathcal{A}$ is defined as $\| \mathcal{A} \|_{F} = \sqrt{\langle \mathcal{A}, \mathcal{A} \rangle}$.  The \textit{Kronecker product} of two matrices $\bm{A} \in \mathbb{R}^{m \times n}$ and $\bm{B} \in \mathbb{R}^{p \times q}$ is an $mp \times nq$ matrix $\bm{A} \otimes \bm{B} = [\bm{a}_1 \otimes \bm{B}\quad  \bm{a}_2 \otimes \bm{B} \quad    \quad \bm{a}_n \otimes \bm{B}]$. The \textit{Khatri-Rao} product of two matrices whose numbers of columns are the same, $\bm{A} \in \mathbb{R}^{m \times r}, \bm{B} \in \mathbb{R}^{p \times r}$, denoted by $\bm{A} \odot \bm{B}$, is a $mp \times r$ matrix defined by $[\bm{a}_{1} \otimes \bm{b}_{1} \quad  \bm{a}_2 \otimes \bm{b}_{2} \quad  \bm{a}_{r} \otimes \bm{b}_{r}]$. If $\bm{a}$ and $\bm{b}$ are vectors, then $\bm{A} \otimes \bm{B} = \bm{A} \odot \bm{B}$.

\section{MPCA and its Challenges in Data Privacy Protection}
\label{s:metho}

In this section, we will first review the multilinear principal component analysis (MPCA) method and its optimization algorithm proposed by \cite{lu2008mpca}. Then, we will discuss the challenges of data privacy protection of the MPCA optimization algorithm under the federated learning context.

\subsection{MPCA and Its Optimization Algorithm }\label{ss: 2.2}

MPCA is a dimension reduction method for tensor objects. It focuses on retaining most of the essential information of high-dimensional tensors while reducing their dimensionality. This is achieved by identifying a set of projection matrices that can effectively project high-dimensional tensors onto a lower-dimensional tensor subspace, with the goal of maximizing the variation of the projected tensors within this subspace

Following the notations in \cite{lu2008mpca}, we consider a data set consists of $M$ tensor objects $\lbrace \mathcal{X}_{m}, m = 1, \ldots, M \rbrace$. Each tensor object $\mathcal{X}_{m} \in \mathbb{R}^{I_1 \times I_2 \times \ldots \times I_N}$ resides in the high-dimensional tensor (multilinear) space $\mathbb{R}^{I_1} \otimes \mathbb{R}^{I_2} \otimes \ldots \otimes \mathbb{R}^{I_N}$, where $\mathbb{R}^{I_1}, \mathbb{R}^{I_2}, \ldots, \mathbb{R}^{I_N}$ are $N$ vector (linear) spaces. The objective of MPCA is to recognize a multilinear transformation  $\lbrace \tilde{\bm{U}}^{(n)} \in \mathbb{R}^{I_n \times P_n}, n = 1, \dots, N \rbrace$ that maps objects in the original high-dimensional tensor space $\mathbb{R}^{I_1} \otimes \mathbb{R}^{I_2} \otimes \ldots \otimes \mathbb{R}^{I_N}$ into a low-dimensional tensor subspace $\mathbb{R}^{P_1} \otimes \mathbb{R}^{P_2} \otimes \ldots \otimes \mathbb{R}^{P_N}$ ($P_n < I_n$ for $n = 1, \dots, N$) so that the variation of the projected tensors is maximized. The projected tensors can be calculated by using $\mathcal{Y}_{m} = \mathcal{X}_{m} \times_{1} \tilde{\bm{U}}^{(1)^\top} \times_{2} \tilde{\bm{U}}^{(2)^\top} \ldots \times_{N} \tilde{\bm{U}}^{(N)^\top}, m = 1, \ldots, M$, where the projected tensors $\mathcal{Y}_{m} \in \mathbb{R}^{P_1 \times P_2 \times \ldots \times P_N}$ are also known as low-dimensional features. To maximize the variation, the following optimization criterion is solved: 

\begin{equation}\label{eq:MPCA 1}
    \lbrace \tilde{\bm{U}}^{(n)}, n = 1, \dots, N \rbrace = \argmax_{\tilde{\bm{U}}^{(1)}, \tilde{\bm{U}}^{(2)}, \ldots, \tilde{\bm{U}}^{(N)}} \Psi_{\mathcal{Y}}.
\end{equation}

\noindent where $\Psi_{\mathcal{Y}}=\sum_{m=1}^{M}\|\mathcal{Y}_m-\bar{\mathcal{Y}}\|_F^2$, and $\bar{\mathcal{Y}} = \frac
{1}{M}\sum_{m=1}^{M}\mathcal{Y}_m$ is the mean of the projected tensors.


Unlike traditional PCA, there is no analytical solution for problem \eqref{eq:MPCA 1}. To solve criterion \eqref{eq:MPCA 1}, \cite{lu2008mpca} proposed an iterative optimization algorithm, which cyclically updates each of the projection matrices in $\lbrace\tilde{\bm{U}}^{(n)}\rbrace_{n=1}^N$. In other words, they solve criterion \eqref{eq:MPCA 1} by iteratively solving $N$ suboptimization problems that optimize $\tilde{\bm{U}}^{(1)}, \tilde{\bm{U}}^{(2)}, \ldots, \tilde{\bm{U}}^{(N)}$, respectively. They have proved that there is an analytical solution for each suboptimization problem. For example, the solution of $\tilde{\bm{U}}^{(n)}$ can be found by calculating the dominant eigenvectors of matrix $\bm{\Phi}^{(n)}$, which is computed from the high-dimensional tensors $\lbrace\mathcal{X}_{m}\rbrace_{m=1}^M$ and other projection matrices $\tilde{\bm{U}}^{(1)},\tilde{\bm{U}}^{(2)},\ldots,\tilde{\bm{U}}^{(n-1)},\tilde{\bm{U}}^{(n+1)},\ldots,\tilde{\bm{U}}^{(N)}$ (see Theorem \ref{the:theorem 1} below).

\begin{theorem}\label{the:theorem 1}
\citep{lu2008mpca} Let $\lbrace \tilde{\bm{U}}^{(n)}, n = 1, \ldots, N \rbrace$ be the solution to (\ref{eq:MPCA 1}). Then, given all the other projection matrices $\tilde{\bm{U}}^{(1)}, \ldots, \tilde{\bm{U}}^{(n-1)}, \tilde{\bm{U}}^{(n+1)}, \ldots, \tilde{\bm{U}}^{(N)}$, the matrix $\tilde{\bm{U}}^{(n)}$ consists of the $P_n$ eignevectors corresponding to the largest $P_n$ eigenvalues of the matrix
\begin{equation}\label{eq:MPCA 2}
    \bm{\Phi}^{(n)} = \sum_{m=1}^{M} (\bm{X}_{m(n)} - \bar{\bm{X}}_{(n)})  \tilde{\bm{U}}_{\bm{\Phi}^{(n)}}  \tilde{\bm{U}}^{\top}_{\bm{\Phi}^{(n)}}  (\bm{X}_{m(n)} - \bar{\bm{X}}_{(n)})^{\top}
\end{equation}

\noindent where $\tilde{\bm{U}}_{\bm{\Phi}^{(n)}} = \big( \tilde{\bm{U}}^{(n+1)} \otimes \tilde{\bm{U}}^{(n+2)} \otimes \ldots \otimes \tilde{\bm{U}}^{(N)} \otimes \tilde{\bm{U}}^{(1)} \otimes \tilde{\bm{U}}^{(2)} \otimes \ldots \otimes \tilde{\bm{U}}^{(n-1)} \big)$; $\bm{X}_{m(n)}$ is the mode-n matricization of tensor $\mathcal{X}_{m}$; $\bar{\bm{X}}_{(n)}$ is the mean of $\lbrace\bm{X}_{m(n)}\rbrace_{m=1}^M$, which is also the mode-n matricization of $\bar{\mathcal{X}}$, the mean of tensors $\lbrace\mathcal{X}_{m}\rbrace_{m=1}^M$.

\end{theorem}

Algorithm 1 summarizes the iterative optimization algorithm for MPCA, which comprises four steps. The first step is \textit{Preprocessing}, which focuses on centering the high-dimensional tensors: $ \tilde{\mathcal{X}}_{m} = \mathcal{X}_{m} - \bar{\mathcal{X}}, m = 1, \ldots, M $, where $\bar{\mathcal{X}} = \frac{1}{M}\sum_{m = 1}^{M}\mathcal{X}_{m}$ is the sample mean. This is necessary since Equation \eqref{eq:MPCA 2} requires data centralization. Let $\tilde{\bm{X}}_{m(n)} $ be the mode-n matricization of the centered tensor $\tilde{\mathcal{X}}_{m}$, Equation \eqref{eq:MPCA 2} can be expressed as $\bm{\Phi}^{(n)} = \sum_{m=1}^{M} \tilde{\bm{X}}_{m(n)}  \tilde{\bm{U}}_{\bm{\Phi}^{(n)}}  \tilde{\bm{U}}^{\top}_{\bm{\Phi}^{(n)}}  \tilde{\bm{X}}_{m(n)}^{\top}$.

The second step is \textit{Initialization}, which is crucial due to the iterative nature of the MPCA algorithm. There are various applicable initialization methods. In \cite{lu2008mpca}, $\tilde{\bm{U}}^{(n)}$ is initialized using the first $P_n$ eigenvectors corresponding to the most significant $P_n$ eigenvalues of the covariance matrix $\bm{\Phi}^{(n)*} = \sum_{m = 1}^{M} \tilde{\bm{X}}_{m(n)}   \tilde{\bm{X}}_{m(n)}^{\top}$, $n = 1, \ldots, N$.

The third step is referred to as \textit{Local Optimization}, where each projection matrix $\tilde{\bm{U}}^{(n)}, n = 1, \ldots, N$, is iteratively optimized in a cyclic manner until either convergence is attained or the maximum iteration number $K$ is reached. In each cycle, the solution of $\tilde{\bm{U}}^{(n)}$ consists of the eigenvectors corresponding to the largest $P_n$ eigenvalues of the matrix $\bm{\Phi}^{(n)}$ defined in Equation (\ref{eq:MPCA 2}) (see Theorem \ref{the:theorem 1}).

The last step is \textit{Projection}, which projects the high-dimensional tensors to the low-dimension subspace using the projection matrices identified by the third step.

\begin{algorithm}[H]\label{alg: MPCA}

\caption{{\fontfamily{qcr}\selectfont
MPCA Optimization Algorithm \citep{lu2008mpca}}}
\textbf{Input:} A set of tensor samples $ \lbrace \mathcal{X}_{m} \in \mathbb{R}^{I_1 \times I_2 \times \ldots \times I_N}, m = 1, \ldots, M \rbrace$, convergence tolerance $\eta$, and the maximum iteration number $K$

\textbf{Output:} Low-dimensional representations $\lbrace \mathcal{Y}_{m} \in \mathbb{R}^{P_1 \times P_2 \times \ldots \times P_N}, m = 1, \ldots, M \rbrace$ of the input tensor samples with maximum variation captured.

\textbf{Algorithm:}

\textbf{Step 1 (Preprocessing):} Center the input samples as  $\lbrace \tilde{\mathcal{X}}_{m} = \mathcal{X}_{m} - \bar{\mathcal{X}}, m = 1, \ldots, M \rbrace$, where $\bar{\mathcal{X}} = \frac{1}{M}\sum_{m = 1}^{M}\mathcal{X}_{m}$ is the sample mean.  

\textbf{Step 2 (Initialization):} Calculate the eigen-decomposition of $\bm{\Phi}^{(n)*} = \sum_{m = 1}^{M} \tilde{\bm{X}}_{m(n)}   \tilde{\bm{X}}_{m(n)}^{T}$ and set $\tilde{\bm{U}}^{(n)}$ to consist of the eigenvectors corresponding to the most significant $P_n$ eigenvalues, for $n = 1, \ldots, N$.

\textbf{Step 3 (Local Optimization):} 

\hspace{3mm} Calculate $\lbrace \tilde{\mathcal{Y}}_m = \tilde{\mathcal{X}}_m \times_{1} \tilde{\bm{U}}^{(1)^\top} \times_{2} \tilde{\bm{U}}^{(2)^\top} \ldots \times_{N} \tilde{\bm{U}}^{(N)^\top},m = 1, 2, \ldots, M \rbrace$.

\hspace{3mm} Calculate $\Psi_{\mathcal{Y}_{0}} = \sum_{m = 1}^{M} \| \tilde{\mathcal{Y}}_m \|_{F}^{2}$ (the mean $\bar{\tilde{\mathcal{Y}}}$ is all zero since $\tilde{\mathcal{X}}_{m}$ is centered).

\For{k = 1 : K}{ 

\For{n = 1 : N}{

* Set the matrix $\tilde{\bm{U}}^{(n)}$ to consist of the $P_n$ eigenvectors of the matrix $\bm{\Phi}^{(n)}$, as defined in Equation (\ref{eq:MPCA 2}), corresponding to the largest $P_n$ eigenvalues.
}

Calculate $\lbrace\tilde{\mathcal{Y}}_{m}, m = 1, \ldots, M \rbrace$ and $\Psi_{\mathcal{Y}_{k}}$.

If $\Psi_{\mathcal{Y}_{k}} - \Psi_{\mathcal{Y}_{k-1}} \le \eta$, break and go to Step 4 }

\textbf{Step 4 (Projection):} The feature tensor after projection is obtained as $\lbrace \mathcal{Y}_m = \mathcal{X}_{m} \times_{1} \tilde{\bm{U}}^{(1)^\top} \times_{2} \tilde{\bm{U}}^{(2)^\top} \ldots \times_{N} \tilde{\bm{U}}^{(N)^\top},m = 1, 2, \ldots, M \rbrace$.

\end{algorithm}

\subsection{The Challenges in the Data Privacy Protection of MPCA}\label{sec: mpcaleakage}

In this subsection, we discuss the challenges in the data privacy protection of the MPCA algorithm proposed by \cite{lu2008mpca} under the context of federated learning. Recall that we assume the existence of a dataset comprising $M$ tensor objects for MPCA. Under federated learning settings, the $M$ samples are from multiple users/participants. We assume that there are $D$ users, and user $d$, $d=1,2,\ldots,D$, has $M_d$ high-dimensional tensor samples, which are denoted as  $\lbrace \mathcal{X}_{m_d}^{d}, \ m_d = 1, \ldots, M_d \rbrace$. For example, $\mathcal{X}_{3}^{1}$ represents the 3rd high-dimensional tensor sample of User 1. It is obvious that $\sum_{d=1}^DM_d=M$. In the context of federated learning, the first three steps of Algorithm 1 raise data privacy leakage concerns, and each will be discussed individually.

\subsubsection{Preprocessing} \label{sec:sec:preprocessing}

The first step is preprocessing, which focuses on data centralization. This requires the computation of the global sample mean using data from all users. Specifically, the sample mean is computed as follows:

\begin{equation}\label{eq:MPCA sample mean}
    \bar{\mathcal{X}} = \frac{\sum_{d=1}^{D}\sum_{m_d=1}^{M_d}\mathcal{X}_{m_d}^{d}}{\sum_{d=1}^{D}M_d}.
\end{equation} 

It can be seen that the computation of $\bar{\mathcal{X}}$ requires the data from all $D$ users to be aggregated, which cannot protect the data privacy of each user. One possible solution is to let each user compute a local mean using its data and share the local mean along with the local sample size with the server on the cloud. Then, the server will take a weighted average of the local means from all users, which yields the global mean. However, this solution still has the data leakage issue since a local mean still contains information of local data, which might not be allowed to be shared with the server.

\subsubsection{Initialization} \label{sec:sec:Initialization}

Recall that the initialization is achieved by computing matrix $\bm{\Phi}^{(n)*}$ and then computing its first $P_n$ eigenvectors. When there are $D$ users, matrix $\bm{\Phi}^{(n)*}$ is calculated as follows:

\begin{equation}\label{eq:phi}
\bm{\Phi}^{(n)*} = \sum_{d=1}^{D}\sum_{m_d=1}^{M_d}\tilde{\bm{X}}_{m_d(n)}^{d} \tilde{\bm{X}}_{m_d(n)}^{d\top}, 
\end{equation}

\noindent where $\tilde{\bm{X}}_{m_d(n)}^{d}$ is the $n$th mode matricization of $\tilde{\mathcal{X}}_{m_d}^{d}$, and $\tilde{\mathcal{X}}_{m_d}^{d}$ is the $m_d$th sample of user $d$ after centralization, i.e., $ \tilde{\mathcal{X}}_{m_d}^{d} = \mathcal{X}_{m_d}^{d} - \bar{\mathcal{X}}$ for $\ m_d = 1, \ldots, M_d, \ d = 1, \ldots, D $. Apparently, Equation \eqref{eq:phi} requires each user to compute a local matrix (i.e., $\sum_{m_d=1}^{M_d}\tilde{\bm{X}}_{m_d(n)}^{d} \tilde{\bm{X}}_{m_d(n)}^{d\top}$) and send it to the server for summation. This cannot protect the data privacy of each user since the local matrix contains the covariance information of each user's data. 

\subsubsection{Local Optimization}\label{sec:sec:Local Optimization}

Similar to the initialization, local optimization also involves the computation of eigenvectors of matrix $\bm{\Phi}^{(n)}$. As shown in Equation \eqref{eq: phi in local optimization}, the computation of $\bm{\Phi}^{(n)}$ requires the addition of matrices from all $D$ users, 

\begin{equation}\label{eq: phi in local optimization}
    \bm{\Phi}^{(n)} = \sum_{d=1}^{D}\sum_{m_d=1}^{M_d} \tilde{\bm{X}}_{m_d(n)}^{d}  \tilde{\bm{U}}_{\bm{\Phi}^{(n)}}  \tilde{\bm{U}}^{\top}_{\bm{\Phi}^{(n)}}  \tilde{\bm{X}}_{m_d(n)}^{d^\top},
\end{equation}

\noindent where $\tilde{\bm{U}}_{\bm{\Phi}^{(n)}} = \big( \tilde{\bm{U}}^{(n+1)} \otimes \tilde{\bm{U}}^{(n+2)} \otimes \ldots \otimes \tilde{\bm{U}}^{(N)} \otimes \tilde{\bm{U}}^{(1)} \otimes \tilde{\bm{U}}^{(2)} \otimes \ldots \otimes \tilde{\bm{U}}^{(n-1)} \big)$. In other words, each user $d$ needs to compute a local matrix $\sum_{m_d=1}^{M_d} \tilde{\bm{X}}_{m_d(n)}^{d}  \tilde{\bm{U}}_{\bm{\Phi}^{(n)}}  \tilde{\bm{U}}^{\top}_{\bm{\Phi}^{(n)}}  \tilde{\bm{X}}_{m_d(n)}^{d^\top}$ and share it with the server. Here, $\tilde{\bm{U}}_{\bm{\Phi}^{(n)}}$ is shared by all users. Although it might be challenging to infer the exact matrix $\tilde{\bm{X}}_{m_d(n)}^{d^\top}$ from the local matrix, the local matrix still provides the information related to the covariance of matrix $\tilde{\bm{X}}_{m_d(n)}^{d^\top}$ and thus cannot be shared with the server. Also, the size of the local matrix might be substantial in applications with tensor objects. Transmitting such large data to the cloud server may result in significant costs, especially given that such transmissions need to occur in a large number of iterations.

To address the aforementioned challenges in the data privacy protection of the MPCA optimization algorithm, we will propose a federated algorithm for MPCA in Section \ref{sec: FMPCA}.

\section{Federated Multilinear Principal Component Analysis}\label{sec: FMPCA}

In this section, we propose a federated multilinear principal component analysis method. Specifically, we will develop three algorithms to address the challenges in data privacy protection of the MPCA optimization algorithm discussed in Sections \ref{sec:sec:preprocessing}, \ref{sec:sec:Initialization}, and \ref{sec:sec:Local Optimization}, respectively.

\subsection{Preprocessing}\label{sec:preprocessing}

Inspired by the practical secure aggregation method proposed by \cite{bonawitz2016practical}, we develop a secure centralization method for MPCA. The secure centralization method works as follows. First, each user computes a local mean using its local data. Second, the local mean is masked by random perturbation. Specifically, each user generates and sends a random tensor to each of the other $D-1$ users. This implies that each user will in total receive $D-1$ randomly generated tensors from other $D-1$ users. Then, each user utilizes the $D-1$ tensors to compute a perturbation tensor and add it to its local mean, which yields a masked local mean. Third, each user sends its masked local mean to the server. Finally, the server computes the global mean using the masked mean tensors from all $D$ users. The proposed secure centralization method protects data privacy because it masks the local means using random perturbations. More importantly, the random perturbations are canceled off when taking the average by the server, which means the proposed method will not compromise the accuracy of the computed global mean. The details of the proposed secure centralization method are discussed below.

Taking user $d$, $d=1,\ldots, D$, as an example, its local mean (denoted as $\bar{\mathcal{X}}_{d}$) can be computed as follows:
\begin{equation*}
    \bar{\mathcal{X}}_{d} = \frac{1}{M_d}\sum_{m_d=1}^{M_d}\mathcal{X}_{m_d}^{d},
\end{equation*}

\noindent where $\mathcal{X}_{m_d}^{d}$ is the $m_d$th sample of user $d$, $m_d = 1, \ldots, M_d$, and $M_d$ is the number of tensor samples user $d$ has. 

To mask the local mean, user $d$ first randomly generates a tensor for each of the other $D-1$ users. We use $\mathcal{S}_{d, d^\prime} \in \mathbb{R}^{I_1 \times I_2 \times \ldots \times I_N}$ to denote the random tensor that user $d$ sends to user $d'$, $d'=1,\ldots, D$, $d'\neq d$. Each entry of $\mathcal{S}_{d, d^\prime}$ can be generated from a random distribution (e.g., a uniform distribution). Then, the following random perturbations are computed:
\begin{equation*}
    \mathcal{R}_{d, d^\prime} = \mathcal{S}_{d, d^\prime} - \mathcal{S}_{d^\prime, d},
\end{equation*}

\noindent where $\mathcal{S}_{d, d^\prime}$ is the random tensor that user $d$ sends to $d'$, and $\mathcal{S}_{d', d}$ is the random tensor that user $d$ receives from user $d'$. With the $D-1$ perturbations, the local mean of user $d$ is masked as follows:
\begin{equation*}
    \bar{\mathcal{X}}_{d}^\prime = \bar{\mathcal{X}}_{d} + \frac{1}{M_d}\sum_{d^\prime =1, d^\prime \neq d}^{D} \mathcal{R}_{d, d^\prime}, d = 1, \ldots, D,
\end{equation*}

\noindent where $\bar{\mathcal{X}}_{d}^\prime$ is the masked sample mean of user $d$. After adding perturbations, user $d$ sends its masked sample mean $\bar{\mathcal{X}}_{d}^\prime$ along with its sample size $M_d$ to the server. The server will compute the global mean by taking a weighted average of the masked sample means from all $D$ users as shown below: 

\begin{equation*}
\bar{\mathcal{X}}=\frac{\sum_{d=1}^{D}M_d \bar{\mathcal{X}}_{d}^{\prime}}{\sum_{d=1}^{D}M_d}.
\end{equation*}

After calculating the global sample mean, each user will download it from the server and centralize its local tensor samples:

\begin{equation*}
    \tilde{\mathcal{X}}_{m_d}^{d} = \mathcal{X}_{m_d}^{d} - \bar{\mathcal{X}}, 
\end{equation*}

\noindent where $\tilde{\mathcal{X}}_{m_d}^{d}$ denotes the $m_d$th centralized tensor sample of user $d$, $m_d = 1, \ldots, M_d$, $ d = 1, \ldots, D$.

Algorithm \ref{alg:federatedcentralization} summarizes the proposed secure centralization method for FMPCA. It can be easily proved that Algorithm \ref{alg:federatedcentralization} computes the global mean of data samples from all $D$ users. The proof can be found in the Appendix. 

\begin{algorithm}[H]\label{alg:federatedcentralization}
\caption{{\fontfamily{qcr}\selectfont
Federated Centralization Algorithm}}


\begin{enumerate}[noitemsep,leftmargin=5mm,label=(\Alph*)]
\item{\textbf{Users} } 

Each user $d$, $d=1,\ldots, D$,
\begin{enumerate}[label=(\alph*),noitemsep]
    \item calculates its local mean: $\Bar{\mathcal{X}_d}=\frac{1}{M_d}\sum_{m_d=1}^{M_d}\mathcal{X}_{m_d}^{d}$,
    \item generates and sends a random tensor $\mathcal{S}_{d,d^{\prime}}$ to user $d'$ and receives a random tensor $\mathcal{S}_{d',d}$ from user $d'$, $\ d^{\prime}=1,\ldots, D$, $d\neq d^{\prime}$,
    \item computes the perturbations  $\mathcal{R}_{d,d^{\prime}}=\mathcal{S}_{d,d^{\prime}}-\mathcal{S}_{d^{\prime},d}$, $d^{\prime}=1,\ldots, D, \ d\neq d^{\prime}$,
    \item masks local mean with perturbations $\Bar{\mathcal{X}}_d^{\prime}=\Bar{\mathcal{X}}_d + \frac{1}{M_d}\sum_{d^{\prime}=1,d^{\prime}\neq d}^{D}\mathcal{R}_{d,d^{\prime}}$,
    \item sends the masked local mean $\Bar{\mathcal{X}}_d^{\prime}$ and sample size $M_d$ to the server.
\end{enumerate}

\item{\textbf{Server} }
\begin{enumerate}[label=(\alph*),noitemsep]
 \item The server computes the global mean $\Bar{\mathcal{X}}=\frac{\sum_{d=1}^{D}M_d \bar{\mathcal{X}}_{d}^{\prime}}{\sum_{d=1}^{D}M_d}$, and then
 \item sends the global mean to users for data centralization.
 \vspace{-3mm}
\end{enumerate}


\end{enumerate}

\end{algorithm}

\subsection{Initialization}\label{sec:initialization}

Recall that the initialization of the MPCA algorithm requires computing matrix $\bm{\Phi}^{(n)*} = \sum_{d=1}^{D}\sum_{m_d=1}^{M_d}\tilde{\bm{X}}_{m_d(n)}^{d} \tilde{\bm{X}}_{m_d(n)}^{d\top}$ and its eigenvectors. Since the computation of matrix $\bm{\Phi}^{(n)*}$ demands each user $d$ to compute a local matrix (i.e., $\sum_{m_d=1}^{M_d}\tilde{\bm{X}}_{m_d(n)}^{d} \tilde{\bm{X}}_{m_d(n)}^{d\top}$) and send it to the server for summation, it fails to protect user data privacy. To address this challenge, we will first prove that the eigenvectors of matrix $\bm{\Phi}^{(n)*}$ are equal to the left singular vectors of a matrix $\bm{A}_{(n)}$ constructed from the data of all users. Next, we will develop a federated initialization algorithm that allows all users to jointly compute the singular vectors of $\bm{A}_{(n)}$ while keeping their data local and confidential. To show that the eigenvectors of matrix $\bm{\Phi}^{(n)*}$ can be computed via SVD, we first give the following proposition.

\begin{prop}\label{pro:prop 1}

The projection matrix $\tilde{\bm{U}}^{(n)}$ in the initialization of MPCA can be equivalently computed by using the following two methods: 

\begin{enumerate}[label=(\alph*),noitemsep]

\item Conducting eigendecomposition on matrix $\bm{\Phi}^{(n)*} = \sum_{d=1}^{D}\sum_{m_d=1}^{M_d}\tilde{\bm{X}}_{m_d(n)}^{d} \tilde{\bm{X}}_{m_d(n)}^{d\top}$, and set the projection matrix $\tilde{\bm{U}}^{(n)}$ to comprise the eigenvectors corresponding to the most significant $P_n$ eigenvalues.

\item Performing SVD on matrix 
\begin{equation*}
\bm{A}_{(n)} = [\tilde{\bm{X}}_{1(n)}^{1}, \ldots, \tilde{\bm{X}}_{M_{1}(n)}^{1}, \tilde{\bm{X}}_{1(n)}^{2}, \ldots, \tilde{\bm{X}}_{M_{2}(n)}^{2}, \ldots, \tilde{\bm{X}}_{1(n)}^{D}, \ldots, \tilde{\bm{X}}_{M_{D}(n)}^{D}],
\end{equation*}

\noindent and set the projection matrix $\tilde{\bm{U}}^{(n)}$ to consist of the left singular vectors corresponding to the most significant $P_n$ singular values, where $\tilde{\bm{X}}_{m_d(n)}^{d}\in \mathbb{R}^{I_n \times (I_1 \times \ldots \times I_{n-1} \times I_{n+1} \times \ldots \times I_N)}$ is the $n$th-mode matricization of user $d$'s $m_d$th tensor sample, $m_d = 1,\ldots, M_d, \ d = 1, \ldots, D$, and $\bm{A}_{(n)} \in \mathbb{R}^{I_n \times (I_1 \times \ldots \times I_{n-1} \times I_{n+1} \times \ldots \times I_N \times \sum_{d = 1}^{D}M_d)}$

\end{enumerate}

\end{prop}

The proof of Proposition \ref{pro:prop 1} can be found in the Appendix. The initialization of the classic MPCA algorithm in \cite{lu2008mpca} employs the method (a) in Proposition \ref{pro:prop 1}. In this article, our federated algorithm is based on method (b). Our algorithm uses the \textit{user separability} characteristic of matrix $\bm{A}_{(n)}$. \textit{User separability} means the columns of matrix $\bm{A}_{(n)}$ are separable among users. To be specific, the first $I_1 \times \ldots \times I_{n-1} \times I_{n+1} \times \ldots \times I_N \times M_1$ columns of matrix $\bm{A}_{(n)}$ are the data of User 1, the subsequent $I_1 \times \ldots \times I_{n-1} \times I_{n+1} \times \ldots \times I_N \times M_2$ columns of data belong to User 2, and so on and so forth. This characteristic provides the possibility of developing a federated incremental updating algorithm that allows each user to use its local data to sequentially update the singular vectors of $\bm{A}_{(n)}$. To be specific, the first user utilizes its data to compute local singular vectors and share them with the second user, who then uses its data to update the singular vectors shared by the first user. The updated singular vectors will then be shared with the next user for updating until all users get involved. As a result, the last user possesses the final singular vectors (i.e., the singular vectors of $\bm{A}_{(n)}$), which will be shared with the server. In the remainder of this subsection, we provide a detailed discussion of the proposed federated updating algorithm.

To perform the proposed federated updating algorithm, users need to preprocess their local data. Specifically, each user $d$ needs to reshape its local data into a flattened matrix, $\bm{A}_{d(n)}$, $d=1\ldots, D$. Recall that the centered data of user $d$ are denoted as $\tilde{\bm{X}}_{m_d(n)}^{d} \in \mathbb{R}^{I_n \times (I_1 \times \ldots \times I_{n-1} \times I_{n+1} \times \ldots \times I_N)}$, $m_d = 1, \ldots, M_d$. The local flatted matrix $\bm{A}_{d(n)}$ is computed by concatenating the mode-$n$ matricization of all $M_d$ tensors of user $d$. That is, $\bm{A}_{d(n)} =[ \tilde{\bm{X}}_{1(n)}^{d}, \tilde{\bm{X}}_{2(n)}^{d},  \ldots,\tilde{\bm{X}}_{M_d(n)}^{d} ] \in \mathbb{R}^{{I_n \times (I_1 \times \ldots \times I_{n-1} \times I_{n+1} \times \ldots \times I_N \times M_d)}}$. For notation simplicity, we use $J_{d(n)}$ to denote the number of columns of $\bm{A}_{d(n)}$, i.e., $J_{d(n)}=I_1 \times \ldots \times I_{n-1} \times I_{n+1} \times \ldots \times I_N \times M_d$. After data preprocessing, the federated updating algorithm works as follows.

First, the first user performs SVD on its local flattened matrix, which yields $\tilde{\bm{A}}_{1(n)}= \tilde{\bm{U}}^{(n)} \bm{\Sigma}^{(n)} \bm{V}^{(n)}$. Then, the left singular matrix (i.e., $\tilde{\bm{U}}^{(n)} \in \mathbb{R}^{I_n \times I_n}$) and the diagonal matrix containing singular values (i.e., $\bm{\Sigma}^{(n)} \in \mathbb{R}^{I_n \times J_{1(n)}}$) are shared with User 2. Matrix $\bm{V}^{(n)}\in \mathbb{R}^{J_{1(n)} \times J_{1(n)} }$ is not shared, so the data privacy of $\tilde{\bm{A}}_{1(n) }$ can be protected. 

Second, User 2 uses its local data to update $\tilde{\bm{U}}^{(n)}$ and $\bm{\Sigma}^{(n)}$ shared by User 1. Recall the flattened matrix of User 2 is $\tilde{\bm{A}}_{2(n)}\in \mathbb{R}^{{I_n \times J_{2(n)}}}$. The most straightforward way to compute the updated singular vectors and singular values is to perform SVD on a concatenated matrix $\tilde{\bm{A}}^{new} = [\tilde{\bm{A}}_{1(n)} \; \tilde{\bm{A}}_{2(n)}]\in \mathbb{R}^{{I_n \times (J_{1(n)}+J_{2(n)})}}$. That is, $\tilde{\bm{A}}^{new} =\tilde{\bm{U}}^{(n)new} \bm{\Sigma}^{(n)new} \bm{V}^{(n)new}$, where $\tilde{\bm{U}}^{(n)new}$ is the updated singular vectors and $\bm{\Sigma}^{(n)new}$ is the updated singular value matrix. However, this is infeasible since the data of the first user (i.e., $\tilde{\bm{A}}_{1(n)}$) cannot be shared with the second user, so it is impossible for User 2 to perform SVD on $\tilde{\bm{A}}^{new}$. To address this challenge, we give the following proposition, which indicates that $\tilde{\bm{U}}^{(n)new}$ can be computed by updating $\tilde{\bm{U}}^{(n)}$ using  $\tilde{\bm{A}}_{2(n)}$.

\begin{prop}\label{pro:update}

Given matrix $\bm{A}\in \mathbb{R}^{m\times n}$ and denote its left singular vector matrix and singular value matrix as $\bm{U}_{A}\in \mathbb{R}^{m\times m}$ and $\bm{\Sigma}_{A}\in \mathbb{R}^{m\times n}$, respectively. Let $\bm{C}=[\bm{A}\;\; \bm{B}]\in \mathbb{R}^{m\times (n+p)}$, where $\bm{B}\in \mathbb{R}^{m\times p}$ is another matrix, and denote the left singular vector matrix and singular value matrix of $\bm{C}$ as $\bm{U}_{C}\in \mathbb{R}^{m\times m}$ and $\bm{\Sigma}_{C}\in \mathbb{R}^{m\times (n+p)}$, respectively. Then, $\bm{U}_{C}$ and $\bm{\Sigma}_{C}$ can be computed by using $\bm{B}$ to update $\bm{U}_{A}$ and $\bm{\Sigma}_{A}$, respectively, following the steps below:

\begin{enumerate}
    \item Project $\bm{B}$ onto the space spanned by the columns of $\bm{U}_{A}$ and compute the residual matrix: $\bm{R}= \bm{B}-\bm{U}_{A}\bm{U}_{A}^{\top}\bm{B}\in \mathbb{R}^{m\times p}$
    \item Perform column-wise normalization on $\bm{R}$ as follows and denote the normalized matrix as $\check{\bm{R}}$:
    \begin{equation*}
     \check{\bm{r}}_j =\begin{cases}
      \frac{\bm{r}_j}{\|\bm{r}_j\|_2} & \text{if $\|\bm{r}_j\|_2\neq 0$}\\
      \bm{r}_j & \text{if $\|\bm{r}_j\|_2= 0$},
    \end{cases}
    \end{equation*}where $\check{\bm{r}}_j$ and ${\bm{r}}_j$ are the $j$th column of $\check{\bm{R}}$ and $\bm{R}$, respectively, $j=1,\ldots,p$. 
    \item Perform SVD on matrix $\bm{M} = \begin{bmatrix} \bm{\Sigma}_{A} & \bm{U}_{A}^{\top}\bm{B} \\ \bm{0} & \bm{E} \end{bmatrix}\in \mathbb{R}^{(m+p)\times (n+p)}$, i.e., $\bm{M} ={{\bm{U}}_{M}}\bm{\Sigma}_{M}\bm{V}_{M}$, where $\bm{0}\in \mathbb{R}^{p\times n}$ is a zero matrix, and $\bm{E}\in \mathbb{R}^{p\times p}$ is a diagonal matrix whose diagonal elements are the $\ell_2$ norm of the columns of $\bm{R}$
    \item Set $\bm{U}_{C}$ be the first $m$ columns of $\Big[\bm{U}_{A} \quad \check{\bm{R}} \Big] \bm{U}_{M}$ and $\bm{\Sigma}_{C}$ be the first $m$ rows of $\bm{\Sigma}_{M}$.

\end{enumerate}
\end{prop}

The proof of Proposition \ref{pro:update} can be found in the Appendix. Proposition \ref{pro:update} suggests that User 2 can use its data $\tilde{\bm{A}}_{2(n)}$ to update the left singular vector matrix and singular value matrix from User 1. Subsequently, these updated singular vectors and values are shared with the third user for further updating, and this process continues until all users get involved. The last user will have the final singular vectors matrix (and singular value matrix) that equals the singular vector matrix (and singular value matrix) of the aggregated matrix containing the data from all users. In the updating process, what is shared among users are the singular vectors and singular values, ensuring the protection of each user's data privacy. Recall that the \textit{Initialization} step of MPCA requires only the first $P_n$ dominant singular vectors, the last user can truncate the singular vector matrix by keeping the first $P_n$ columns and share the truncated matrix with all other users. We summarize the federated initialization algorithm in Algorithm 3.

\begin{algorithm}\label{alg:federatedinitialization}
\caption{{\fontfamily{qcr}\selectfont
Federated Initialization Algorithm
}}
\textbf{Input:} D users and their local demeaned tensor data, $\tilde{\mathcal{X}}_{m_d}^{d} \in \mathbb{R}^{I_1 \times I_2 \times \ldots \times I_N}$, $\ m_d = 1, \ldots, M_d$, $d = 1, \ldots, D $.

\textbf{Output:} Projection matrix $\tilde{\bm{U}}^{(n)} \in \mathbb{R}^{I_n \times P_n}$ for the initialization of MPCA

\textbf{Preprocessing:} Each user $d$ unfolds its tensor data into $n$th mode matrices, $\lbrace \tilde{\bm{X}}_{m_d(n)}^{d} \in \mathbb{R}^{I_n \times (I_1 \times \ldots \times I_{n-1} \times I_{n+1} \times \ldots \times I_N)}\rbrace_{m_d=1}^{M_d}$, and then concatenates them into a matrix $\bm{A}_{d(n)} = [ \tilde{\bm{X}}_{1(n)}^{d}, \tilde{\bm{X}}_{2(n)}^{d},  \ldots,\tilde{\bm{X}}_{M_d(n)}^{d} ] \in \mathbb{R}^{{I_n \times J_{d(n)}}}$


\For{\textbf{User} $d = 1 : D $}{

\uIf{$d==1$} 
{
\textit{\%The first user}

Conduct SVD on $\tilde{\bm{A}}_{1(n)}$, i.e., $\tilde{\bm{A}}_{1(n)}= \tilde{\bm{U}}^{(n)} \bm{\Sigma}^{(n)} \bm{V}^{(n)}$

Share $\tilde{\bm{U}}^{(n)}\in \mathbb{R}^{I_n\times I_n}$ and $\bm{\Sigma}^{(n)}\in \mathbb{R}^{{I_n \times J_{1(n)}}} $ with User 2
}
\Else{


Project data $\tilde{\bm{A}}_{d(n)}$ onto the space spanned by the columns of $\tilde{\bm{U}}^{(n)}$ and compute the residual matrix $\bm{R}=\tilde{\bm{A}}_{d(n)}-\tilde{\bm{U}}^{(n)}\tilde{\bm{U}}^{(n)\top}\tilde{\bm{A}}_{d(n)}$

Construct a new matrix $\bm{M}=\begin{bmatrix}
\bm{\Sigma}^{(n)} & \tilde{\bm{U}}^{(n)\top}\tilde{\bm{A}}_{d(n)}\\
\bm{0} & \bm{E} \end{bmatrix}$, where $\bm{E}$ is a diagonal matrix whose diagonal elements are the $\ell_2$ norm of the columns of $\bm{R}$\\

Conduct SVD on $\bm{M}$, i.e., $\bm{M}=\bm{U}\bm{\Sigma}\bm{V}$\vspace{2mm}

Set $\tilde{\bm{U}}^{(n)}$ be the first $I_n$ columns of $[\tilde{\bm{U}}^{(n)}\;\; \check{\bm{R}}]\bm{U}$ and $\bm{\Sigma}^{(n)}$ be the first $I_n$ rows of $\bm{\Sigma}$, where $\check{\bm{R}}$ is normalized from $\bm{R}$ (see Step 2 of Proposition \ref{pro:update} for details)  

\uIf{$d \neq D$} 
{
Share $\tilde{\bm{U}}^{(n)}$ and $\bm{\Sigma}$ with the next user

}
\Else{
\textit{\%The last user}

Truncate $\tilde{\bm{U}}^{(n)}$ by keeping its first $P_n$ columns and share it with all other users
}
}}
\end{algorithm}

\newpage

\normalsize

\subsection{Local Optimization}\label{sec:localoptimization}


Recall that the local optimization involves the computation of eigenvectors of matrix $\bm{\Phi}^{(n)}$, which is computed from the matrices of all $D$ users: 

\begin{equation*}\label{eq: phi in local optimization2}
    \bm{\Phi}^{(n)} = \sum_{d=1}^{D}\sum_{m_d=1}^{M_d} \tilde{\bm{X}}_{m_d(n)}^{d}  \tilde{\bm{U}}_{\bm{\Phi}^{(n)}}  \tilde{\bm{U}}^{\top}_{\bm{\Phi}^{(n)}}  \tilde{\bm{X}}_{m_d(n)}^{d^\top},
\end{equation*}

\noindent where $\tilde{\bm{U}}_{\bm{\Phi}^{(n)}} = \big( \tilde{\bm{U}}^{(n+1)} \otimes \tilde{\bm{U}}^{(n+2)} \otimes \ldots \otimes \tilde{\bm{U}}^{(N)} \otimes \tilde{\bm{U}}^{(1)} \otimes \tilde{\bm{U}}^{(2)} \otimes \ldots \otimes \tilde{\bm{U}}^{(n-1)} \big)$. Section \ref{sec:initialization} discussed the federated initialization algorithm that provides the projection matrices $ \tilde{\bm{U}}^{(n)}, n = 1, \ldots, N$, to all users. Thus, all users have access to matrix $\tilde{\bm{U}}_{\bm{\Phi}^{(n)}}$. However, $\bm{\Phi}^{(n)}$ requires each user $d$ to compute a local matrix $\sum_{m_d=1}^{M_d} \tilde{\bm{X}}_{m_d(n)}^{d}  \tilde{\bm{U}}_{\bm{\Phi}^{(n)}}  \tilde{\bm{U}}^{\top}_{\bm{\Phi}^{(n)}}  \tilde{\bm{X}}_{m_d(n)}^{d^\top}$ and share it with the server, which cannot protect user data privacy and can be computationally intensive (see Section \ref{sec: mpcaleakage} for more details). To address these challenges, we will first show that the eigenvectors of matrix $\bm{\Phi}^{(n)}$ equal the left singular vectors of a matrix $\bm{A}_{(n)}^{\bm{\Phi}}$, which is constructed from the data of all users. Next, we will show that matrix $\bm{A}_{(n)}^{\bm{\Phi}}$ also has the \textit{user separability} characteristic, which enables us to develop a federated incremental updating algorithm that allows multiple users to jointly compute its left singular vectors while keeping each user's data local and confidential.

\begin{prop}\label{pro:localoptimization2}

The projection matrix $\tilde{\bm{U}}^{(n)}$ in the local optimization step of MPCA can be equivalently computed by using the following two methods: 

\begin{enumerate}[label=(\alph*),noitemsep]

\item Conducting eigendecomposition on matrix $\bm{\Phi}^{(n)} = \sum_{d=1}^{D}\sum_{m_d=1}^{M_d} \tilde{\bm{X}}_{m_d(n)}^{d}  \tilde{\bm{U}}_{\bm{\Phi}^{(n)}}  \tilde{\bm{U}}^{\top}_{\bm{\Phi}^{(n)}}  \tilde{\bm{X}}_{m_d(n)}^{d^\top}$, and set the projection matrix $\tilde{\bm{U}}^{(n)}$ to comprise the eigenvectors corresponding to the most significant $P_n$ eigenvalues.

\item Performing SVD on matrix 
\begin{equation*}
\bm{A}_{(n)}^{\bm{\Phi}} = [\tilde{\bm{X}}_{1(n)}^{1\bm{\Phi}}, \ldots, \tilde{\bm{X}}_{M_{1}(n)}^{1\bm{\Phi}}, \tilde{\bm{X}}_{1(n)}^{2\bm{\Phi}}, \ldots, \tilde{\bm{X}}_{M_{2}(n)}^{2\bm{\Phi}}, \ldots, \tilde{\bm{X}}_{1(n)}^{D\bm{\Phi}}, \ldots, \tilde{\bm{X}}_{M_{D}(n)}^{D\bm{\Phi}}],
\end{equation*}

\noindent and set the projection matrix $\tilde{\bm{U}}^{(n)}$ to consist of the left singular vectors corresponding to the most significant $P_n$ singular values, where $ \tilde{\bm{X}}_{m_d(n)}^{d\bm{\Phi}} = \tilde{\bm{X}}_{m_d(n)}^{d} \tilde{\bm{U}}_{\bm{\Phi}^{(n)}}\in \mathbb{R}^{I_n \times (P_1 \times \ldots \times P_{n-1} \times P_{n+1} \times \ldots \times P_N)}$, $m_d = 1,\ldots, M_d, \ d = 1, \ldots, D$, \\and $\bm{A}_{(n)}^{\bm{\Phi}} \in \mathbb{R}^{I_n \times (P_1 \times \ldots \times P_{n-1} \times P_{n+1} \times \ldots \times P_N \times \sum_{d=1}^{D}M_d)}$

\end{enumerate}

\end{prop}

\begin{algorithm}\label{alg:federatedlocaloptimization}
\caption{{\fontfamily{qcr}\selectfont
Federated Local Optimization Algorithm}}
\textbf{Input:} D users and their local demeaned tensor data, $\tilde{\mathcal{X}}_{m_d}^{d} \in \mathbb{R}^{I_1 \times I_2 \times \ldots \times I_N}$, $\ m_d = 1, \ldots, M_d$, $d = 1, \ldots, D $, and the projection matrices $\tilde{\bm{U}}^{(n)}, n=1,\ldots, N$ from Algorithm \ref{alg:federatedinitialization}

\textbf{Output:} Projection matrix $\tilde{\bm{U}}^{(n)} \in \mathbb{R}^{I_n \times P_n}$

\textbf{Preprocessing:} Each user $d$ first unfolds its tensor data into $n$th mode matrices, $\lbrace \tilde{\bm{X}}_{m_d(n)}^{d} \in \mathbb{R}^{I_n \times (I_1 \times \ldots \times I_{n-1} \times I_{n+1} \times \ldots \times I_N)}\rbrace_{m_d=1}^{M_d}$. Next, each user $d$ computes $\lbrace \tilde{\bm{X}}_{m_d(n)}^{d\bm{\Phi}} = \tilde{\bm{X}}_{m_d(n)}^{d} \tilde{\bm{U}}_{\bm{\Phi}^{(n)}}\rbrace_{m_d=1}^{M_d}$, where $\tilde{\bm{U}}_{\bm{\Phi}^{(n)}} = \big( \tilde{\bm{U}}^{(n+1)} \otimes \tilde{\bm{U}}^{(n+2)} \otimes \ldots \otimes \tilde{\bm{U}}^{(N)} \otimes \tilde{\bm{U}}^{(1)} \otimes \tilde{\bm{U}}^{(2)} \otimes \ldots \otimes \tilde{\bm{U}}^{(n-1)} \big)$. Finally, each user $d$ concatenates its local data into a matrix $\bm{A}_{d(n)}^{\bm{\Phi}} = [\tilde{\bm{X}}_{1(n)}^{d\bm{\Phi}}, \ldots, \tilde{\bm{X}}_{M_{d}(n)}^{d\bm{\Phi}}]\in \mathbb{R}^{I_n \times (P_1 \times \ldots \times P_{n-1} \times P_{n+1} \times \ldots \times P_N\times M_d)}$.
 

\For{\textbf{User} $d = 1 : D $}{

\uIf{$d==1$} 
{
\textit{\%The first user}

Conduct SVD on $\bm{A}_{1(n)}^{\bm{\Phi}}$, i.e., $\bm{A}_{1(n)}^{\bm{\Phi}}= \tilde{\bm{U}}^{(n)} \bm{\Sigma}^{(n)} \bm{V}^{(n)}$

Share $\tilde{\bm{U}}^{(n)}\in \mathbb{R}^{I_n\times I_n}$ and $\bm{\Sigma}^{(n)}\in \mathbb{R}^{{I_n \times K_{1(n)}}} $ with User 2
}
\Else{


Project data $\bm{A}_{1(n)}^{\bm{\Phi}}$ onto the space spanned by the columns of $\tilde{\bm{U}}^{(n)}$ and compute the residual matrix $\bm{R}=\bm{A}_{1(n)}^{\bm{\Phi}}-\tilde{\bm{U}}^{(n)}\tilde{\bm{U}}^{(n)\top}\bm{A}_{1(n)}^{\bm{\Phi}}$

Construct a new matrix $\bm{M}=\begin{bmatrix}
\bm{\Sigma}^{(n)} & \tilde{\bm{U}}^{(n)\top}\bm{A}_{1(n)}^{\bm{\Phi}}\\
\bm{0} & \bm{E} \end{bmatrix}$, where $\bm{E}$ is a diagonal matrix whose diagonal elements are the $\ell_2$ norm of the columns of $\bm{R}$\\

Conduct SVD on $\bm{M}$, i.e., $\bm{M}=\bm{U}\bm{\Sigma}\bm{V}$\vspace{2mm}

Set $\tilde{\bm{U}}^{(n)}$ be the first $I_n$ columns of $[\tilde{\bm{U}}^{(n)}\;\; \check{\bm{R}}]\bm{U}$ and $\bm{\Sigma}^{(n)}$ be the first $I_n$ rows of $\bm{\Sigma}$, where $\check{\bm{R}}$ is normalized from $\bm{R}$ (see Step 2 of Proposition 3 for details)  

\uIf{$d \neq D$} 
{
Share $\tilde{\bm{U}}^{(n)}$ and $\bm{\Sigma}$ with the next user

}
\Else{
\textit{\%The last user}

Truncate $\tilde{\bm{U}}^{(n)}$ by keeping its first $P_n$ columns and share it with all other users
}
}}
\end{algorithm}

The proof of Proposition \ref{pro:localoptimization2} can be found in the Appendix. The MPCA algorithm in \cite{lu2008mpca} employs the method (a) in Proposition \ref{pro:localoptimization2}. In this article, our federated algorithm is based on method (b). It is easy to show that matrix $\bm{A}_{(n)}^{\bm{\Phi}}$ also has the \textit{user separability} characteristic, meaning the columns of matrix $\bm{A}_{(n)}^{\bm{\Phi}}$ are separable among users. To be specific, the first $P_1 \times \ldots \times P_{n-1} \times P_{n+1} \times \ldots \times P_N \times M_1$ columns of matrix $\bm{A}_{(n)}^{\bm{\Phi}}$ are the data of User 1, the subsequent $P_1 \times \ldots \times P_{n-1} \times P_{n+1} \times \ldots \times P_N \times M_2$ columns of data belong to User 2, and so on and so forth. This characteristic provides the possibility of developing a federated incremental updating algorithm that allows each user to use its local data to sequentially update the singular vectors of $\bm{A}_{(n)}^{\bm{\Phi}}$. Similar to the federated initialization algorithm discussed in Section \ref{sec:initialization}, the first user utilizes its data to compute local singular vectors and share them with the second user, who then uses its data to update the singular vectors shared by the first user. The updated singular vectors will then be shared with the next user for updating until all users get involved. As a result, the last user possesses the final singular vectors (i.e., the singular vectors of $\bm{A}_{(n)}^{\bm{\Phi}}$), which will be shared with other users. We provide a detailed step-by-step discussion of the proposed federated updating algorithm below.

First, each user preprocesses its local data. Each user $d$ constructs a concatenated matrix $\bm{A}_{d(n)}^{\bm{\Phi}} = [\tilde{\bm{X}}_{1(n)}^{d\bm{\Phi}}, \ldots, \tilde{\bm{X}}_{M_{d}(n)}^{d\bm{\Phi}}]\in \mathbb{R}^{I_n \times (P_1 \times \ldots \times P_{n-1} \times P_{n+1} \times \ldots \times P_N\times M_d)}$, $d=1, \ldots, D$, where $ \tilde{\bm{X}}_{m_d(n)}^{d\bm{\Phi}} = \tilde{\bm{X}}_{m_d(n)}^{d} \tilde{\bm{U}}_{\bm{\Phi}^{(n)}}$. For notation simplicity, we use $K_{d(n)}$ to denote the number of columns of $\bm{A}_{d(n)}^{\bm{\Phi}}$, i.e., $K_{d(n)}=P_1 \times \ldots \times P_{n-1} \times P_{n+1} \times \ldots \times P_N \times M_d$. Second, the first user performs SVD on its local matrix, which yields $\bm{A}_{1(n)}^{\bm{\Phi}}= \tilde{\bm{U}}^{(n)} \bm{\Sigma}^{(n)} \bm{V}^{(n)}$. Then, the left singular matrix (i.e., $\tilde{\bm{U}}^{(n)} \in \mathbb{R}^{I_n \times I_n}$) and the diagonal matrix containing singular values (i.e., $\bm{\Sigma}^{(n)} \in \mathbb{R}^{I_n \times K_{1(n)}}$) are shared with User 2. Matrix $\bm{V}^{(n)}\in \mathbb{R}^{K_{1(n)} \times K_{1(n)} }$ is not shared, so the data privacy of $\bm{A}_{1(n)}^{\bm{\Phi}}$ can be protected. Third, User 2 uses its local data to update $\tilde{\bm{U}}^{(n)}$ and $\bm{\Sigma}^{(n)}$. Specifically, User 2 uses its data $\tilde{\bm{A}}_{2(n)}$ to update the left singular vector matrix and singular value matrix from User 1 based on Proposition \ref{pro:update}. Subsequently, these updated singular vectors and values are shared with the third user for further updating, and this process continues until all users get involved. The last user will have the final singular vectors matrix (and singular value matrix) that equals the singular vector matrix (and singular value matrix) of the aggregated matrix containing the data from all users. In the updating process, what is shared among users are the singular vectors and singular values, ensuring the protection of each user's data privacy. Recall that the \textit{Local Optimization} step of MPCA requires only the first $P_n$ dominant singular vectors, the last user can truncate the singular vector matrix by keeping the first $P_n$ columns and sharing the truncated matrix with all other users. We summarize the federated local optimization algorithm in Algorithm 3.

\subsection{The Federated MPCA Algorithm}\label{sss: 2.4.4}

In this subsection, we summarize the proposed federated multilinear principal component analysis method. Recall there are $D$ users, and each user has $M_d$ high-dimensional tensor samples $\lbrace \mathcal{X}_{m_d}^{d} \in \mathbb{R}^{I_1 \times I_2 \times \ldots \times I_N}, m_d = 1, \ldots, M_d, d = 1, \ldots, D \rbrace$. In the first step, \textit{Preprocessing}, all users jointly centralize their data by using {\fontfamily{qcr}\selectfont Algorithm \ref{alg:federatedcentralization}: Federated Centralization}, which yields the centered data $\lbrace \tilde{\mathcal{X}}_{m_d}^{d} \in \mathbb{R}^{I_1 \times I_2 \times \ldots \times I_N}, m_d = 1, \ldots, M_d, d = 1, \ldots, D \rbrace$.

In the second step, \textit{Initilization}, all users jointly initialize $\lbrace\tilde{\bm{U}}^{(n)}\rbrace_{n=1}^N$ using {\fontfamily{qcr}\selectfont Algorithm \ref{alg:federatedinitialization}:Federated Initialization}. Since Algorithm \ref{alg:federatedinitialization} is an incremental algorithm that allows each user to use its own data to update the projection matrix from the former user, the last user possesses the final initialization matrices $\lbrace\tilde{\bm{U}}^{(n)}\rbrace_{n=1}^N$, which are then sent to the server to be shared with all users.

In the third step, \textit{Local Optimization}, all users first jointly calculate the initial scatter. Specifically, each user $d$ downloads the initialized projection matrices from the server and computes its local low-dimensional tensors: $\tilde{\mathcal{Y}}_{m_d}^{d} = \tilde{\mathcal{X}}_{m_d}^{d} \times_{1} \tilde{\bm{U}}^{(1)^\top} \times_{2} \tilde{\bm{U}}^{(2)^\top} \ldots \times_{N} \tilde{\bm{U}}^{(N)^\top}$, $m_d = 1, \ldots, M_d, \ d = 1, \ldots, D$. Then, user $d$ calculates its local scatter $\sum_{m_d = 1}^{M_d}\| \tilde{\mathcal{Y}}_{m_d}^{d}\|_{F}^{2}$, which is shared with the server. The server aggregates the local scatters from all $D$ users ${\Psi}_{\mathcal{Y}_{0}}=\sum_{d = 1}^{D}\sum_{m_d = 1}^{M_d}\|\tilde{\mathcal{Y}}_{m_d}^{d}\|_{F}^{2}$, which is the initial scatter that will be used to determine the convergence of the algorithm. 

Then, the iterative updating of projection matrices $\{\tilde{\bm{U}}^{(n)}\}_{n=1}^N$ starts. The algorithm allows a maximum of $K$ iterations, where the value $K$ can be set by the server. In each iteration $k$, $k=1,\ldots, K$, the $N$ projection matrices $\tilde{\bm{U}}^{(1)},\tilde{\bm{U}}^{(2)},\ldots,\tilde{\bm{U}}^{(N)}$ are updated one by one. Taking matrix $\tilde{\bm{U}}^{(n)},n=1,\ldots, N,$ as an example, each user first downloads $\lbrace\tilde{\bm{U}}^{(j)}\rbrace_{j=1,j\neq n}^N$ from the server (this is not necessary for the first iteration since the projection matrices have been downloaded when calculating the initial scatter). After that, all users jointly update $\tilde{\bm{U}}^{(n)}$ using {\fontfamily{qcr}\selectfont Algorithm \ref{alg:federatedlocaloptimization}:Federated Local Optimization}, and the last user sends the updated $\tilde{\bm{U}}^{(n)}$ to the server. Once the updating of all $N$ projection matrices is complete, the server possesses the latest updated $\lbrace\tilde{\bm{U}}^{(n)}\rbrace_{n=1}^N$. Then, the server needs to check if the FMPCA algorithm has converged or if the maximum number of iterations $K$ has been met or not. To check if the convergence of the algorithm is achieved, each user $d$ downloads $\lbrace\tilde{\bm{U}}^{(n)}\rbrace_{n=1}^N$ from the server and calculates $\sum_{m_d = 1}^{M_d}\| \tilde{\mathcal{Y}}_{m_d}^{d}\|_{F}^{2}$ and send it to the server, where $\tilde{\mathcal{Y}}_{m_d}^{d} = \tilde{\mathcal{X}}_{m_d}^{d} \times_{1} \tilde{\bm{U}}^{(1)^\top} \times_{2} \tilde{\bm{U}}^{(2)^\top} \ldots \times_{N} \tilde{\bm{U}}^{(N)^\top}$, and the server calculates the updated scatter
${\Psi}_{\mathcal{Y}_{k}}=\sum_{d = 1}^{D}\sum_{m_d = 1}^{M_d}\|\tilde{\mathcal{Y}}_{m_d}^{d}\|_{F}^{2}$. If $\Psi_{\mathcal{Y}_{k}} - \Psi_{\mathcal{Y}_{k-1}} \le \eta$ is satisfied, it implies the algorithm has converged.  If the algorithm has converged or the maximum number of iterations $K$ is achieved, the \textit{Local Optimization} is completed. 

The last step is \textit{Projection}, which yields the low-dimensional features of MPCA. In this step, each user $d$ downloads $\lbrace\tilde{\bm{U}}^{(n)}\rbrace_{n=1}^N$ from the server and computes its local low-dimensional tensors
$\lbrace \mathcal{Y}_{m_d}^{} = \mathcal{X}_{m_d}^{d} \times_{1} \tilde{\bm{U}}^{(1)^\top} \times_{2} \tilde{\bm{U}}^{(2)^\top} \ldots \times_{N} \bm{U}^{(N)^\top}\rbrace_{m_d = 1}^{M_d}$. The proposed Federated MPCA method is summarized in Algorithm \ref{alg: FMPCA}.

\begin{algorithm}[H]\label{alg: FMPCA}
\caption{{\fontfamily{qcr}\selectfont
Federated MPCA}}
\textbf{Input:} A set of D users where each user has $ {M}_d$ high-dimensional tensor samples $ \mathcal{X}_{m_d}^{d} \in \mathbb{R}^{I_1 \times I_2 \times \ldots \times I_N}, m_d = 1, \ldots, M_d, \ d = 1, \ldots, D$, the convergence tolerance $\eta$, and the maximum iteration number $K$

\textbf{Output:} Low-dimensional tensor features of each user $ \mathcal{Y}_{m_d}^{d} \in \mathbb{R}^{P_1 \times P_2 \times \ldots \times P_N},$ $ m_d = 1, \ldots, M_d, \ d = 1, \ldots, D$.

\textbf{Step 1 (Preprocessing):}  All users jointly centralize $\lbrace\lbrace\mathcal{X}_{m_d}^{d}\rbrace_{ m_d = 1}^{M_d}\rbrace_{d=1}^D$ to get $\lbrace\lbrace\tilde{\mathcal{X}}_{m_d}^{d}\rbrace_{ m_d = 1}^{M_d}\rbrace_{d=1}^D$  using {\fontfamily{qcr}\selectfont Algorithm \ref{alg:federatedcentralization}:Federated Centralization}.

\textbf{Step 2 (Initialization):}   All users jointly initialize $\lbrace\tilde{\bm{U}}^{(n)}\rbrace_{n=1}^N$ using {\fontfamily{qcr}\selectfont Algorithm \ref{alg:federatedinitialization}:Federated Initialization}, and the last user sends $\lbrace\tilde{\bm{U}}^{(n)}\rbrace_{n=1}^N$ to the server.

\textbf{Step 3 (Local Optimization):}

Each user $d$ downloads $\lbrace\tilde{\bm{U}}^{(n)}\rbrace_{n=1}^N$ from the server and calculates $\sum_{m_d = 1}^{M_d}\| \tilde{\mathcal{Y}}_{m_d}^{d}\|_{F}^{2}$ and send it to the server, where $\tilde{\mathcal{Y}}_{m_d}^{d} = \tilde{\mathcal{X}}_{m_d}^{d} \times_{1} \tilde{\bm{U}}^{(1)^\top} \times_{2} \tilde{\bm{U}}^{(2)^\top} \ldots \times_{N} \tilde{\bm{U}}^{(N)^\top}$

The server calculates
${\Psi}_{\mathcal{Y}_{0}}=\sum_{d = 1}^{D}\sum_{m_d = 1}^{M_d}\|\tilde{\mathcal{Y}}_{m_d}^{d}\|_{F}^{2}$

\For{k = 1 : K}{

\For{n = 1 : N}{
Each user downloads $\lbrace\tilde{\bm{U}}^{(j)}\rbrace_{j=1,j\neq n}^N$ from the server

All users jointly update $\tilde{\bm{U}}^{(n)}$ using {\fontfamily{qcr}\selectfont Algorithm \ref{alg:federatedlocaloptimization}:Federated Local Optimization}, and the last user sends the updated $\tilde{\bm{U}}^{(n)}$ to the server
}

Each user $d$ downloads $\lbrace\tilde{\bm{U}}^{(n)}\rbrace_{n=1}^N$ from server and calculates $\sum_{m_d = 1}^{M_d}\| \tilde{\mathcal{Y}}_{m_d}^{d}\|_{F}^{2}$ and send it to the server, where $\tilde{\mathcal{Y}}_{m_d}^{d} = \tilde{\mathcal{X}}_{m_d}^{d} \times_{1} \tilde{\bm{U}}^{(1)^\top} \times_{2} \tilde{\bm{U}}^{(2)^\top} \ldots \times_{N} \tilde{\bm{U}}^{(N)^\top}$

The server calculates
${\Psi}_{\mathcal{Y}_{k}}=\sum_{d = 1}^{D}\sum_{m_d = 1}^{M_d}\|\tilde{\mathcal{Y}}_{m_d}^{d}\|_{F}^{2}$

The server checks if $\Psi_{\mathcal{Y}_{k}} - \Psi_{\mathcal{Y}_{k-1}} \le \eta$ or $k == K$ is satisfied. If yes, break and go to \textbf{Step 4} 
}

$\textbf{Step 4 (Projection):} $  

Each user $d$ downloads $\lbrace\tilde{\bm{U}}^{(n)}\rbrace_{n=1}^N$ from server and computes its local low-dimensional features
$\lbrace \mathcal{Y}_{m_d}^{} = \mathcal{X}_{m_d}^{d} \times_{1} \tilde{\bm{U}}^{(1)^\top} \times_{2} \tilde{\bm{U}}^{(2)^\top} \ldots \times_{N} \bm{U}^{(N)^\top}\rbrace_{m_d = 1}^{M_d}$

\end{algorithm}

\section{The Application of FMPCA in Industrial Prognostics}\label{sec:prognostics}

In this section, we explore the application of the proposed FMPCA method in the field of industrial prognostic, which focuses on leveraging the degradation signal of engineering assets to predict their failure times. The failure of industrial assets typically results from a gradual and irreversible damage accumulation process known as degradation. Although a degradation process is typically not directly observable, it is often associated with some indicators that can be observed and captured through sensing technology, which yields degradation signals. Industrial prognostic works by establishing a data analytics model that maps assets' degradation signals to their failure times. Similar to numerous other data analytics models, the development of an industrial prognostic model typically comprises two stages: model training and model testing/utilization. Model training involves using historical data (often referred to as training data), which comprises degradation signals and failure times from some failed assets, to train the model. Once the model has been trained, meaning that its parameters have been estimated, the second stage is to monitor the condition of an in-field asset to obtain its real-time degradation signal and input the degradation signal into the trained model to predict the asset's failure time.

The training of prognostic models requires an appropriate amount of historical data to achieve a desirable performance. However, the number of failed assets (and thus the amount of historical data) from a single organization might be very limited. Consequently, a single organization might not be able to train a reliable and effective prognostic model on its own. Thus, it is beneficial to allow multiple organizations to jointly train a prognostic model together using their data. However, the data from each organization usually cannot be merged or shared with others due to privacy constraints. To address this challenge, there have been some efforts to develop federated industrial prognostic models \citep{su2022federated}, which allow multiple users to collaboratively train a prognostic model while keeping each user's data local and confidential. However, existing federated prognostic models are designed for time series-based degradation signals but not imaging data. Imaging-based degradation signals have been used more and more commonly in recent years in the prediction of TTFs \citep{fang2019image}. This trend is driven by the noncontact nature of imaging sensing technology, making it more feasible for implementation. Furthermore, it offers a more comprehensive set of information compared to degradation signals based on time series data. Imaging data usually exhibits complex spatiotemporal correlation structures -- that is-- the pixels within each image have spatial correlation, and pixels at the same location across multiple images are temporally correlated. To capture such complex correction structures, imaging data are often modeled as tensors. As imaging data are typically ultra-high dimensional, tensor-based dimension reduction methods such as MPCA are commonly utilized first to extract low-dimensional features, and these low-dimensional features are then employed in the construction of prognostic models.

In this article, we apply the proposed FMPCA method to develop a federated imaging data-based prognostic model. We consider the scenario that multiple users possess imaging-based degradation signals from identical assets. However, the amount of data each user has is too limited to independently train a reliable prognostic model, and the data from these users cannot be simply shared and merged due to privacy concerns. The model consists of two steps: dimension reduction and prognostic model construction. In the first step, the proposed FMPCA method is employed by these users to jointly reduce the dimension of their imaging data, producing low-dimensional features. Subsequently, a prognostic model is established by regressing the low-dimensional features against the assets' TTFs using the federated (log)-location-scale regression \citep{su2022federated}. 


Following the notations in Section \ref{sss: 2.4.4}, we denote the training high-dimensional image-based degradation data as $ \mathcal{X}_{m_d}^{d} \in \mathbb{R}^{I_1 \times I_2 \times I_3}, m_d = 1, \ldots, M_d, d = 1, \ldots, D $, where $D$ is the number of users, $M_d$ is the number of failed assets (i.e., samples) that user $d$ has. Also, the TTFs corresponding to these failed assets are denoted as $\tilde{z}_{m_d}^{d} \in \mathbb{R}, m_d = 1,\ldots, M_d, d = 1, \ldots, D$. The proposed FMPCA method (i.e., Algorithm \ref{alg: FMPCA}) is first applied by all $D$ users to jointly perform dimension reduction. This yields the projection matrices $ \tilde{\bm{U}}^{(1)} \in \mathbb{R}^{I_1 \times P_1},\tilde{\bm{U}}^{(2)} \in \mathbb{R}^{I_2 \times P_2}$, $\tilde{\bm{U}}^{(3)} \in \mathbb{R}^{I_3 \times P_3}$ that are shared by all users and each user can compute its low-dimensional features locally:

\begin{equation}\label{eq:training}
    {\mathcal{Y}}_{m_d}^{d} = \mathcal{X}_{m_d}^{d} \times_{1} \tilde{\bm{U}}^{(1)^\top} \times_{2} \tilde{\bm{U}}^{(2)^\top}\times_{3} \tilde{\bm{U}}^{(3)^\top}, \  m_d = 1, \ldots, M_d, \ d = 1, \ldots, D.
\end{equation}

Next, a federated prognostic model is jointly trained by all users. Recall that each user $d,\ d = 1, \ldots, D $, has the low-dimensional features of its training assets (i.e., $ {\mathcal{Y}}_{m_d}^{d}, \  m_d = 1, \ldots, M_d$) and their TTFs (i.e., $\tilde{z}^d_{m_d}, \ m_d = 1, \ldots, M_d$). We will employ the federated algorithm proposed by \cite{su2022federated} to allow $D$ users to jointly train the following LLS regression model while keeping each user's features and TTFs local and confidential: 

\begin{equation}\label{eq:pred t}
    z_m = \beta_0 + vec({\mathcal{Y}}_m)^\top \bm{\beta}_{1} + \sigma\epsilon_{m},
\end{equation}

\noindent where $m=1,\ldots,M$ and $M = \sum_{d = 1}^{D}M_d$ is the total number of training samples all $D$ users have. $z_m$ represents the TTF of asset $m$ if a location-scale regression model is selected. Alternatively, if a log-location-scale regression model is chosen, $z_m$ corresponds to the logarithm of the TTF of asset $m$. $vec({\mathcal{Y}}_m)\in \mathbb{R}^{(P_1 \times P_2 \times P_3) \times 1}$ is the vectorization of ${\mathcal{Y}}_{m}$, $\beta_0 \in \mathbb{R}$ and $\bm{\beta}_{1} \in \mathbb{R}^{(P_1 \times P_2 \times P_3) \times 1}$ are the regression coefficients, $\sigma$ is the scale parameter, and $\epsilon_{i}$ is the random noise with a standard location-scale probability density function $f(\epsilon)$. For example, $f(\epsilon) = 1/\sqrt{2\pi}\exp(-\epsilon^{2}/2)$ for a normal distribution and $f(\epsilon) = \exp(\epsilon - \exp(\epsilon))$ for an SEV distribution. We denote the estimated parameters as $\hat\beta_0,\hat{\bm{\beta}_{1}}$, and $\hat{\sigma}$, which are shared by all $D$ users.  

After the joint training of the prognostic model, each user may use the model to monitor the condition of its in-filed assets and predict their TTFs in real time. Take user $d$ as an example and denote the real-time degradation signal of one of its assets as $\mathcal{X}_{t}^d\in \mathbb{R}^{I_1 \times I_2 \times I_3}$, the low-dimensional features can be extracted as follows:

\begin{equation}
    {\mathcal{Y}}_{t}^{d} = \mathcal{X}_{t}^{d} \times_{1} \tilde{\bm{U}}^{(1)^\top} \times_{2} \tilde{\bm{U}}^{(2)^\top}\times_{3} \tilde{\bm{U}}^{(3)^\top}
\end{equation}

\noindent where the projection matrices $ \tilde{\bm{U}}^{(1)} \in \mathbb{R}^{I_1 \times P_1},\tilde{\bm{U}}^{(2)} \in \mathbb{R}^{I_2 \times P_2}$, $\tilde{\bm{U}}^{(3)} \in \mathbb{R}^{I_3 \times P_3}$ are from the training (same as the ones in Equation \eqref{eq:training}). The low-dimensional feature ${\mathcal{Y}}_{t}^{d}$ is then plugged into the trained LLS regression to predict the asset's failure time distribution: $\hat{z}_{t}^d \sim LLS(\hat{\beta}_{0} + vec({\mathcal{Y}}_{t}^d)^\top\hat{\bm{\beta}}_{1}, \hat{\sigma})$, where $\hat{\beta}_{0} + vec({\mathcal{Y}}_{t}^d)^\top\hat{\bm{\beta}}_{1}$ and $\hat{\sigma}$ are respectively the estimated location and scale parameters.

One challenge in real-time industrial prognostics is that the number of degradation images varies across different assets due to failure truncation. Specifically, an asset is taken out of operation for maintenance or replacement upon failure. This implies that degradation data can only be observed until the failure time, and no additional degradation data can be captured beyond that point. Also, the failure times of different assets are usually different. Hence, the number of degradation images for various assets in the training dataset typically varies. Also, the number of images a test asset has is also typically different from that of the assets in the training dataset since the number of images a test asset has is increasing over time. To address this challenge, we may employ the time-varying framework that has been widely used in industrial prognostics \citep{muller2005time, fang2015adaptive}. To be specific, assume that the length of the test asset is $I_t$--that is--$\mathcal{X}_{t}^d\in \mathbb{R}^{I_1 \times I_2 \times I_t}$, then all users truncate their training assets by keeping only the first $I_t$ images of each asset; if the number of degradation images an asset has is less than $I_t$, then this asset is excluded from the training dataset. By doing so, we can guarantee that the training data and the test data have the same number of images such that the aforementioned federated industrial prognostic model can be applied. To expedite real-time failure time prediction, a strategy is to initially train multiple prognostic models offline using the first $1$, $2$, $3,\ldots$ images. Subsequently, the model trained with the same number of images as the test data is selected for TTF prediction.

\section{Numerical Study}\label{sec:sim}

In this section, we validate the effectiveness of our proposed FMPCA as well as its application in industrial prognostics using synthetic data.

\subsection{Data Generation} 
The simulated data generation is based on a heat transfer process. Specifically, the image stream of asset $m$, which is denoted by $\mathcal{X}_{m}(x, y, t), m = 1, 2, \ldots, 500$, is generated using the following formula:

\begin{equation*}
\frac{\partial \mathcal{X}_{m}(x,y,t) }{\partial t} = 
\alpha_{m} \Big( \frac{\partial^{2} \mathcal{X}_{m}}{\partial x^{2}} + \frac{\partial^{2} \mathcal{X}_{m}}{\partial y^{2}}\Big),
\end{equation*}
    
\noindent where $ (x,y), 0 \le x,y \le 0.2$, represents the location of each image pixel. $\alpha_{m}$ is the thermal diffusivity coefficient for asset $m$, and is randomly generated from a uniform distribution $\mathcal{U}(0.5 \times 10^{-4}, 1 \times 10^{-4})$. $t$ is the time frame. The initial and boundary conditions are set such that $\mathcal{X}|_{t = 1} = 0 $ and $\mathcal{X}_m|_{x = 0} = \mathcal{X}_m|_{x = 0.2} = \mathcal{X}_m|_{y = 0} = \mathcal{X}_m|_{y = 0.2} = 30$. At each time $t$, the image is recorded at locations $x = \frac{j}{n + 1}, y = \frac{k}{n + 1}, j, k = 1, \ldots, n$, resulting in an $n \times n $ matrix. Here, we set $n = 21$ and $t = 1, 2, \dots, 150$, which generates 150 images of size 21 $\times$ 21 for each asset. For the convenience of computation complexity, we select the images whose indices are $15, 30, \ldots, 150$ which leads to 10 images of size 21 $\times$ 21 for each asset represented by 21 $\times$ 21 $\times$ 10 tensor. Next, an independent and identically noise $\epsilon \sim N(0, 0.1)$ is added to each pixel. In total, we generate degradation image streams for 500 assets.

To simulate the TTF of each asset, we first apply MPCA to the generated image streams, which provides projection matrices $\bm{U}^{(1)}\in \mathbb{R}^{21\times P_1}, \bm{U}^{(2)}\in\mathbb{R}^{21\times P_2}, \bm{U}^{(3)}\in\mathbb{R}^{10\times P_3}$,  where the values $P_1, P_2, P_3$ are determined by setting the variation kept in each mode to 97\% as suggested by \cite{lu2008mpca}. Then, an IID random noise from $\mathcal{N}(0,1)$ is added to each entry of the projection matrices, which yields $\tilde{\bm{U}}^{(1)}$, $\tilde{\bm{U}}^{(2)}$, and $\tilde{\bm{U}}^{(3)}$. Next, the low-dimensional features are computed by using $\mathcal{Y}_{m} = \mathcal{X}_{m} \times_{1} \tilde{\bm{U}}^{(1)\top} \times_{2} \tilde{\bm{U}}^{(2)\top}\times_{3}\tilde{\bm{U}}^{(3)\top}, m = 1, \ldots, M$. Each element of coefficients $\beta_{0} \in \mathbb{R}$ and $\bm{\beta}_{1} \in \mathbb{R}^{(P_1 \times P_2 \times P_3) \times 1}$ are generated from a  normal distribution $\mathcal{N}(0,0.01)$. Finally, the TTFs are generated from $\log(z_{m}) = \beta_{0} + \textit{vec}(\mathcal{Y}_{m})^\top\bm{\beta}_{1} + \epsilon_{m}, m = 1, \ldots, 500$, where the IID random noise $\epsilon_{m} \sim \bm{N}(0, 0.1)$.

\subsection{Benchmarks and Performance Evaluation Metric}

We randomly choose 400 assets to create a training dataset and reserve the remaining 100 assets for testing. Among the 400 assets in the training dataset, 250 assets are assigned to User 1, 100 assets to User 2, and the remaining 50 assets to User 3.

We compare the performance of our proposed method (designated as ``FMPCA") with 4 benchmarks. The first benchmark, denoted as ``User Combination", is a non-federated model that utilizes all the 400 assets in the training data set for model training. Specifically, MPCA is first applied to the degradation images of all 400 assets to extract low-dimensional features. Then, a lognormal regression is established by regressing the $400$ TTFs against the low-dimensional features. The other three benchmarks are individually trained models, meaning each user trains one model using its own data. Specifically, the second benchmark (labeled as ``User 1") is identical to the first benchmark, with the distinction that it only utilizes the 250 assets from User 1 for model training. Similarly, the third benchmark (referred to as ``User 2") and the fourth benchmark model (referred to as ``User 3") are analogous to the second benchmark, except that they employ the 100 assets from User 2 and the 50 assets from User 3 to train the prognostic model, respectively. The dimension of low-dimensional features $P_1 \times P_2 \times P_3$ of the proposed FMPCA and four benchmarks is determined by 10-fold cross-validation. 

After the training of the proposed FMPCA-based model and four benchmarks, the degradation images from the $100$ test assets are used to evaluate the prediction performance of these models. In other words, the proposed method as well as the benchmarks are validated using the same test dataset. The prediction errors are calculated from the following equation:

\begin{equation}\label{eq:error}
\text{Prediction Error} = \frac{\vert \text{Estimated TTF} - \text{True TTF} \vert}{\text{True TTF}}.
\end{equation}

We replicate the entire evaluation process, starting from the random splitting of the 500 assets into training and test datasets, 10 times. The prediction errors are reported in Figure \ref{fig: Numerical Study}.

\subsection{Results and analysis} 

\begin{figure*}[!htp]
\centering
 \includegraphics[width=0.8\textwidth]{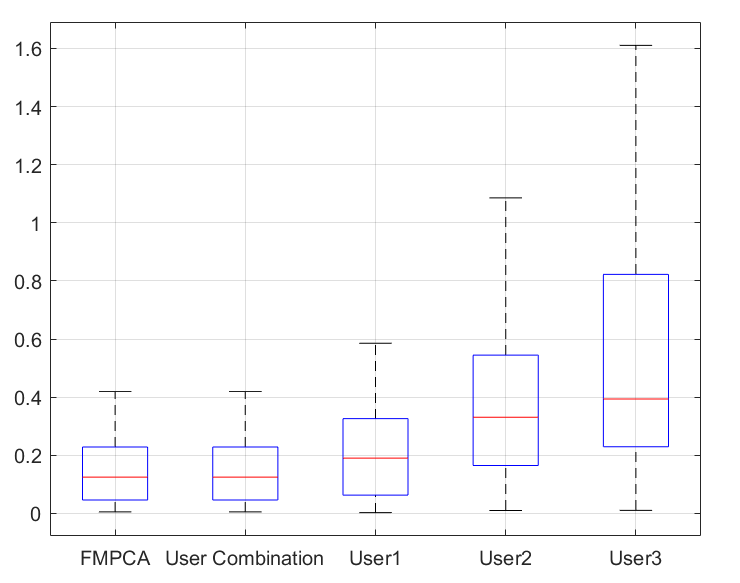}
 \caption{Numerical Study.}
 \label{fig: Numerical Study}
\end{figure*}

Figure \ref{fig: Numerical Study} illustrates that the performance of our proposed method (``FMPCA") is the same as the first benchmark ``User Combination". For example, the medians, first quartile (Q1), and third quartile (Q3) of the absolute prediction errors of ``FMPCA" are $0.13$, $0.03$, and $0.21$, respectively, which are identical to those of ``User Combination". We also checked the projection matrices $\tilde{\bm{U}}^{(1)}, \tilde{\bm{U}}^{(2)}, \tilde{\bm{U}}^{(3)}$ estimated by our proposed method, and noticed that they are exactly the same as the projection matrices from the first benchmark, which uses classic MPCA \citep{lu2008mpca}. This result is expected, as we have demonstrated that the functionality of the three proposed algorithms—{\fontfamily{qcr}\selectfont Federated Centralization Algorithm}, {\fontfamily{qcr}\selectfont Federated Initialization Algorithm}, and {\fontfamily{qcr}\selectfont Federated Local Optimization Algorithm}—mirrors that of the three steps (i.e., \textit{Preprocessing}, \textit{Initialization}, and \textit{Local Optimization}) in the classic MPCA algorithm. Importantly, these federated algorithms maintain the data privacy of each user. In other words, our proposed FMPCA method allows multiple users to jointly perform MPCA while keeping each user's data local and confidential. In the meanwhile, it can achieve the same performance as applying MPCA on a dataset aggregated from all users. 

Figure \ref{fig: Numerical Study} also shows the performance of the proposed method outperforms that of the three benchmarks (i.e., ``User 1", ``User 2" and `` User 3"), where each user trains a prognostic model using its data. For example, the median (and the Interquartile Range, i.e., IQRs) of the absolute prediction errors for ``FMPCA" and ``User 1", ``User 2", ``User 3" are 0.13 (0.17), 0.19 (0.28), 0.37 (0.39), 0.40 (0.59), respectively. This suggests that individual users can benefit by participating in federated learning. Figure \ref{fig: Numerical Study} also indicates that ``User 1" performs the best, while ``User 3" exhibits the least favorable performance among the three individually trained benchmarks. Recall that the training sample sizes for "User 1," "User 2," and "User 3" are $250$, $100$, and $50$, respectively. This demonstrates how the training sample size plays a crucial role in influencing the performance of data analytic models, especially when the training sample size is small. This once again affirms the advantages of participating in federated learning, particularly for users with very limited data size for model training.

\section{Case Study} \label{sec:case}
In this section, we validate the effectiveness of our proposed method by using degradation image streams obtained from a rotating machinery test bed. The experimental test bed is designed to perform accelerated degradation tests on rolling element thrust bearings. The test bearings are run from brand new to failure. Degradation images are collected by an FLIR T300 infrared camera. In the meanwhile, an accelerometer was mounted on the test bed to monitor the vibration of the bearing. Failure time is defined once the amplitude of the vibration signal crosses a threshold based on ISO standards for machine vibration. Each infrared image has 40 $\times $ 20 pixels. In total 284 degradation image streams and their corresponding TTFs are generated. More details about the data set can be found in \cite{gebraeel2009residual} and \cite{fang2019image}.

\begin{figure*}[!htp]
\centering
 \includegraphics[width=0.7\textwidth]{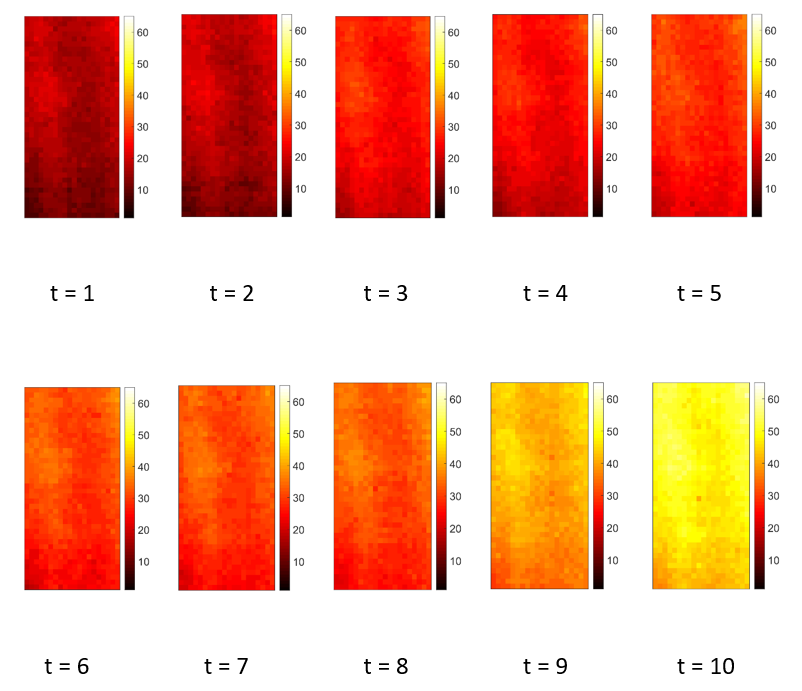}
 \caption{An illustration of one infrared degradation image stream. }
 \label{fig: Case Study_infrared}
\end{figure*}

\begin{figure*}[!htp]
\centering
 \includegraphics[width=0.8\textwidth]{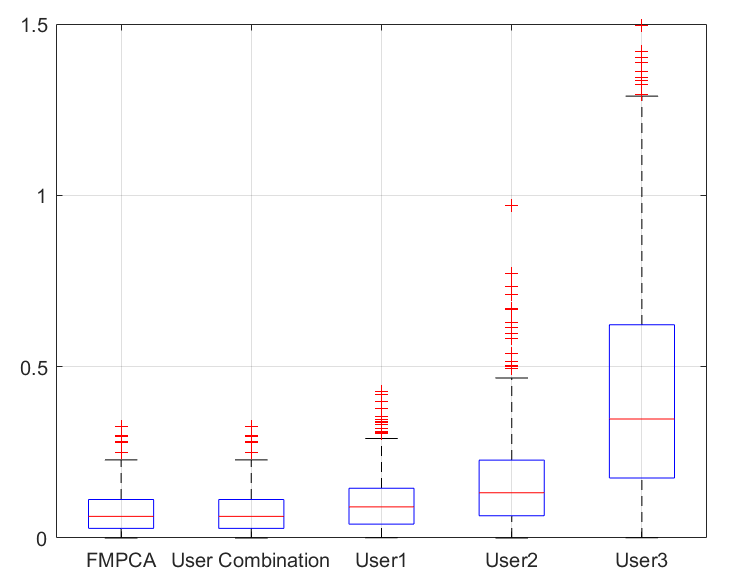}
 \caption{Case Study.}
 \label{fig: Case Study}
\end{figure*}

As discussed in detail in Section \ref{sec:prognostics}, different assets may have different numbers of images due to failure time truncation. To address this challenge, one may employ the time-varying framework that has been widely used in industrial prognostics \citep{muller2005time, fang2015adaptive}. This varying length issue can also be seen as missing data problem, where the data of the asset with the longest TTF in the dataset is considered complete, and the data of other assets are considered to have missing images. In this case study, we first apply a tensor completion method known as TMac developed by \cite{xu2013parallel} to impute the missing images. Compared to using the time-varying framework, tensor completion helps reduce the computational time. More importantly, it will not impact the evaluation of the proposed FMPCA method and the benchmarks because both the proposed method and the benchmarking models are assessed using the same imputed dataset. Similar to the numerical study, we randomly split the 284 imputed image streams into a training data set containing 227 assets and a test data set consisting of the remaining 57 assets. The ratio is roughly 80\% for training and 20\% for testing. In the training data set, 140 assets are randomly assigned to User 1, 57 assets to User 2, and the remaining 30 assets to User 3. Similar to the numerical study, we use 10-fold cross-validation to determine the dimension of the features. In addition, we compare the performance of the proposed ``FMPCA" method with the four benchmarks used in Section \ref{sec:sim}, i.e., ``User Combined", ``User 1", ``User 2", and ``User 3". The same test dataset (containing the data from the $57$ assets) is used to evaluate the prediction performance and the prediction errors are computed using Equation \eqref{eq:error}. We repeat the whole process 10 times and report the results in Figure \ref{fig: Case Study}. 

Figure \ref{fig: Case Study} indicates that our proposed method can achieve the same prediction performance as the first benchmark ``User Combination". For example, the medians, Q1s, and Q3s of the absolute prediction errors of both methods are 0.06, 0.02, and 0.1, respectively. 
Additionally, we compare the projection matrices $\tilde{\bm{U}}^{(1)}, \tilde{\bm{U}}^{(2)}, \tilde{\bm{U}}^{(3)}$ from ``FMPCA" with those from ``User Combined" and observe that they are identical. These confirmed that the proposed FMPCA method has the same performance as the classic MPCA method \citep{lu2008mpca}. The distinction lies in the fact that classic MPCA necessitates the combination of data from different users before its application. However, the proposed FMPCA allows for the implementation of MPCA using data from multiple users while maintaining the privacy and confidentiality of each user's data.

Figure \ref{fig: Case Study} further illustrates that the predictive performance of the proposed ``FMPCA" method surpasses that of the other three benchmarks - ``User 1", ``User 2", and ``User 3". For example, the median (and IQR) of the absolute prediction errors of ``User Combination" and ``User 1", ``User 2", and ``User 3" are $0.06 (0.08)$, $0.1 (0.11)$, $0.15 (0.2)$, and $0.31 (0.47)$, respectively. This again confirms the benefits of participating in federated learning, especially for users with limited training sample size. For example, User 3 has training data from only $30$ assets and the median (and IQR) of the absolute prediction errors from the model trained by itself is $0.31 (0.47)$. However, if User 3 gets involved in the federated learning method, the median (and IQR) of the absolute prediction errors of its prognostic model is $0.06  (0.08)$. Also, it can be seen from Figure \ref{fig: Case Study} that, among the three benchmarks, ``User 1" demonstrates the highest performance, followed by ``User 2," and ``User 3" exhibits the least favorable prediction performance. Recalling that the sample sizes for ``User 1," ``User 2," and ``User 3" are $140$, $57$, and $30$, respectively, this once again emphasizes the importance of having a sufficient number of samples for model training and also the advantages of the proposed FMPCA.

\section{Conclusions}\label{sec:conclusion}

Tensors are increasingly employed in machine learning and data analysis due to their capability to model complex correlation structures in multi-dimensional data. However, given that tensors are typically ultra-high dimensional, methods for tensor dimension reduction are often applied to reduce their dimension. Among these methods, multilinear principal component analysis (MPCA) \citep{lu2008mpca} stands out as the most widely used dimensionality reduction technique for tensor data. However, there has been no exploration of integrating MPCA with federated learning, an important technique that allows a model to be trained across multiple decentralized devices holding local data samples. To address this gap, this paper proposed a federated multilinear principal component analysis (FMPCA) method, which enables multiple users to collaboratively perform MPCA to reduce the dimension of their tensor data while keeping each user's data local and confidential. This is achieved by developing three federated learning algorithms—{\fontfamily{qcr}\selectfont Federated Centralization Algorithm}, {\fontfamily{qcr}\selectfont Federated Initialization Algorithm}, and {\fontfamily{qcr}\selectfont Federated Local Optimization Algorithm}—to replace the three key steps (i.e., \textit{Preprocessing}, \textit{Initialization}, and \textit{Local Optimization}) in the traditional MPCA method. The performance of the proposed FMPCA method is guaranteed to be equivalent to that of traditional MPCA.

As an illustration of the applications of the proposed FMPCA, we implemented it in image data-based industrial prognostics, which focuses on using imaging-based degradation signals to predict the time-to-failure of engineering assets. Using a simulated dataset as well as a data set from rotating machinery, we evaluated the effectiveness of the proposed FMPCA method as well as the industrial prognostic model. The results verified that FMPCA achieves the same performance as that of traditional MPCA. In addition, the results illustrated that the predictive performance of the federated learning model surpasses that of individual models trained by each user. This confirms the importance of participating in federated learning, particularly for a user with limited training data unable to construct a reliable and effective model independently.

\if1\blind{
\section*{Acknowledgements}
The authors acknowledge the generous support from the funding agency of XYZ.	} \fi

\newpage

~~~~~~~~~~~~~~~~~~~~~~~~~~~~~~~~~~~~~~~~~~~~~{\LARGE\textbf{Appendix}}
\setcounter{section}{0}

\section{Proof of the Federated Centralization Algorithm}

Recall that $\mathcal{R}_{d, d^\prime} = \mathcal{S}_{d, d^\prime} - \mathcal{S}_{d^\prime, d}$ and $\bar{\mathcal{X}}_{d}^\prime = \bar{\mathcal{X}}_{d} + \frac{1}{M_d}\sum_{d^\prime =1, d^\prime \neq d}^{D} \mathcal{R}_{d, d^\prime}, d = 1, \ldots, D$, we have the following:

\begin{equation*}
\begin{split}
    \frac{\sum_{d=1}^{D}M_d \bar{\mathcal{X}}_{d}^{\prime}}{\sum_{d=1}^{D}M_d}&=\frac{\sum_{d=1}^{D}M_d (\bar{\mathcal{X}}_{d} + \frac{1}{M_d}\sum_{d^\prime=1,d^\prime \neq d}^{D}\mathcal{R}_{d,d^\prime}) }{\sum_{d=1}^{D}M_d}\\ &=\frac{\sum_{d=1}^{D}M_d \bar{\mathcal{X}}_{d} + \sum_{d=1}^{D}\sum_{d^\prime=1,d^\prime \neq d}^{D}\mathcal{R}_{d,d^\prime} }{\sum_{d=1}^{D}M_d}\\
    &=\frac{\sum_{d=1}^{D}M_d \bar{\mathcal{X}}_{d} + \sum_{d=1}^{D}\sum_{d^\prime=1,d^\prime \neq d}^{D}\mathcal{S}_{d,d^{\prime}}-\sum_{d=1}^{D}\sum_{d^\prime=1,d^\prime \neq d}^{D}\mathcal{S}_{d^{\prime},d} }{\sum_{d=1}^{D}M_d}\\
    &=\frac{\sum_{d=1}^{D}M_d \bar{\mathcal{X}}_{d} }{\sum_{d=1}^{D}M_d}\\
    &=\bar{\mathcal{X}}
    \end{split}
\end{equation*}

\section{Proof of Proposition 1}

Recall that $\bm{A}_{(n)} = [\tilde{\bm{X}}_{1(n)}^{1}, \ldots, \tilde{\bm{X}}_{M_{1}(n)}^{1}, \tilde{\bm{X}}_{1(n)}^{2}, \ldots, \tilde{\bm{X}}_{M_{2}(n)}^{2}, \ldots, \tilde{\bm{X}}_{1(n)}^{D}, \ldots, \tilde{\bm{X}}_{M_{D}(n)}^{D}]$, where $\lbrace \tilde{\bm{X}}_{m_d(n)}^{d} \in \mathbb{R}^{I_n \times (I_1 \times \ldots \ldots \times I_{n-1} \times I_{n+1} \times \ldots \times I_N)}, \ m_d = 1, \ldots, M_d, \ d = 1, \ldots, D \rbrace$ are concatenated horizontally in $\bm{A}_{(n)}$. Thus, $\bm{A}_{(n)}\bm{A}_{(n)}^\top = \sum_{d=1}^{D}\sum_{m_d=1}^{M_d}\tilde{\bm{X}}_{m_d(n)}^{d} \tilde{\bm{X}}_{m_d(n)}^{d\top}= \bm{\Phi}^{(n)*}$. Applying singular value decomposition on $\bm{A}_{(n)}$ yields $\bm{A}_{(n)} = \bm{U}^{(n)}\bm{\Sigma}\bm{V}^\top$, where $\bm{U}^{(n)}$ is the left unitary matrix, $\bm{\Sigma}$ is a diagonal matrix whose diagonal entries are the singular values, and $\bm{V}$ is the right unitary matrix. Therefore, $\bm{\Phi}^{(n)*}=\bm{A}_{(n)}\bm{A}_{(n)}^\top = \bm{U}^{(n)}\bm{\Sigma}\bm{V}^\top (\bm{U}^{(n)}\bm{\Sigma}\bm{V}^\top)^\top = \bm{U}^{(n)}\bm{\Sigma}\bm{V}^\top \bm{V}\bm{\Sigma}\bm{U}^{(n)\top}= \bm{U}^{(n)}\bm{\Sigma}^{2}\bm{U}^{(n)\top}$ since $\bm{V}^\top\bm{V}=\bm{I}$. It is known that $\bm{U}^{(n)\top}\bm{U}^{(n)}=\bm{I}$, and $\bm{\Sigma}^{2}$ is a diagonal matrix whose diagonal entries are nonnegative. Thus, the left unitary matrix $\bm{U}^{(n)}$ from the singular value decomposition of $\bm{A}_{(n)}$ is equivalent to the eigenvector matrix of $\bm{\Phi}^{(n)*}$.

\section{Proof of Proposition 2}

Recall that $\check{\bm{R}}\in \mathbb{R}^{m\times p}$ is the column-wise normalized ${\bm{R}}\in \mathbb{R}^{m\times p}$, which implies the $j$th column of $\check{\bm{R}}$ is either $\frac{\bm{r}_j}{\|\bm{r}_j\|_2}$ (when $\|\bm{r}_j\|_2\neq 0$) or a zero vector $\bm{r}_j$ (when $\|\bm{r}_j\|_2= 0$). Also, recall that $\bm{E}\in \mathbb{R}^{p\times p}$ is a diagonal matrix whose diagonal elements are the $\ell_2$ norm of the columns of $\bm{R}$:
 \begin{equation*}
 \bm{E}=
  \begin{bmatrix}
    \|\bm{r}_1\|_2 & & \\
    & \ddots & \\
    & & \|\bm{r}_p\|_2
  \end{bmatrix}
 \end{equation*}

\noindent Therefore, we have $\bm{R}=\check{\bm{R}}\bm{E}$. Also, recall that $\bm{R}= \bm{B}-\bm{U}_{A}\bm{U}_{A}^{\top}\bm{B}$, thus we have $\bm{B}=\bm{U}_{A}\bm{U}_{A}^{\top}\bm{B}+\check{\bm{R}}\bm{E}$. In addition, $\bm{A}=\bm{U}_A\bm{\Sigma}_A\bm{V}_A$, and $\bm{M} = \begin{bmatrix} \bm{\Sigma}_{A} & \bm{U}_{A}^{\top}\bm{B} \\ \bm{0} & \bm{E} \end{bmatrix} ={{\bm{U}}_{M}}\bm{\Sigma}_{M}\bm{V}_{M}$, we have
\begin{equation*}
\begin{split}
\underset{m\times (n+p) }{\bm{C}}&=\begin{bmatrix}\underset{m\times n}{\bm{A}}\;\; \underset{m\times p}{\bm{B}}\end{bmatrix}\\
&=\begin{bmatrix}\underset{m\times m}{\bm{U}_A}\underset{m\times n}{\bm{\Sigma}_A}\underset{n\times n}{\bm{V}_A}\;\; \underset{m\times p}{\bm{B}}\end{bmatrix}\\
&=\begin{bmatrix}\underset{m\times m}{\bm{U}_A} \quad \underset{m\times p}{\check{\bm{R}}} \end{bmatrix}\begin{bmatrix} \underset{m\times n}{\bm{\Sigma}_A} & \underset{m\times p}{\bm{U}_{A}^{\top}\bm{B}} \\ \underset{p\times n}{\bm{0}} & \underset{p\times p}{\bm{E}} \end{bmatrix} \begin{bmatrix}
\underset{n\times n}{\bm{V}_A} & \underset{n\times p}{\bm{0}} \\
\underset{p\times n}{\bm{0}} & \underset{p\times p}{\bm{1}} \end{bmatrix}\\
&=\begin{bmatrix}\underset{m\times m}{\bm{U}_A} \quad \underset{m\times p}{\check{\bm{R}}} \end{bmatrix}\underset{(m+p)\times (m+p)}{\bm{U}_{M}}\underset{(m+p)\times (n+p) }{\bm{\Sigma}_{M}}\underset{(n+p)\times (n+p) }{\bm{V}_{M}}\begin{bmatrix}
\underset{n\times n}{\bm{V}_A} & \underset{n\times p}{\bm{0}} \\
\underset{p\times n}{\bm{0}} & \underset{p\times p}{\bm{1}} \end{bmatrix}\\
&=\underset{m\times(m+p) }{\bm{U}}\underset{(m+p)\times (n+p)}{\bm{\Sigma}}\underset{(n+p)\times (n+p)}{\bm{V}}
\end{split}
\end{equation*}

\noindent Thus, we have $\bm{U}=\begin{bmatrix}\bm{U}_{A} \quad \check{\bm{R}} \end{bmatrix}{{\bm{U}}_{M}}$, $\bm{\Sigma}=\bm{\Sigma}_{M}$, and $\bm{V}=\bm{V}_{M}\begin{bmatrix}
\bm{V}_A & \bm{0} \\
\bm{0} & \bm{1} \end{bmatrix}$. Performing SVD on matrix $\bm{C}$, we have $\bm{C}=\bm{U}_C\bm{\Sigma}_C\bm{V}_C$. Therefore, $\bm{U}_C\in\mathbb{R}^{m\times m}$ is the first $m$ column of $\bm{U}$, and $\bm{\Sigma}_C$ is the first $m$ rows of $\bm{\Sigma}$.

\section{Proof of Proposition 3}

Recall that $\bm{A}_{(n)}^{\bm{\Phi}} = [\tilde{\bm{X}}_{1(n)}^{1\bm{\Phi}}, \ldots, \tilde{\bm{X}}_{M_{1}(n)}^{1\bm{\Phi}}, \tilde{\bm{X}}_{1(n)}^{2\bm{\Phi}}, \ldots, \tilde{\bm{X}}_{M_{2}(n)}^{2\bm{\Phi}}, \ldots, \tilde{\bm{X}}_{1(n)}^{D\bm{\Phi}}, \ldots, \tilde{\bm{X}}_{M_{D}(n)}^{D\bm{\Phi}}]$, where $ \tilde{\bm{X}}_{m_d(n)}^{d\bm{\Phi}} = \tilde{\bm{X}}_{m_d(n)}^{d} \tilde{\bm{U}}_{\bm{\Phi}^{(n)}}\in \mathbb{R}^{I_n \times (P_1 \times \ldots \times P_{n-1} \times P_{n+1} \times \ldots \times P_N)}$, $m_d = 1,\ldots, M_d, \ d = 1, \ldots, D$ are concatenated horizontally to form $\bm{A}_{(n)}^{\bm{\Phi}}$. Thus, we have the following $\bm{A}_{(n)}^{\bm{\Phi}}\bm{A}_{(n)}^{\bm{\Phi}^\top} = \sum_{d=1}^{D}\sum_{m_d =1 }^{M_d} \sum_{d=1}^{D}\sum_{m_d=1}^{M_d} \tilde{\bm{X}}_{m_d(n)}^{d}\tilde{\bm{U}}_{\bm{\Phi}^{(n)}}  \tilde{\bm{U}}^{\top}_{\bm{\Phi}^{(n)}}  \tilde{\bm{X}}_{m_d(n)}^{d\top}= \bm{\Phi}^{(n)}$. Applying singular value decomposition on $\bm{A}_{(n)}^{\bm{\Phi}}$ yields $\bm{A}_{(n)}^{\bm{\Phi}} = \bm{U}^{(n)}\bm{\Sigma}\bm{V}^\top$, where $\bm{U}^{(n)}$ is the left unitary matrix, $\bm{\Sigma}$ is a diagonal matrix whose diagonal entries are the singular values, and $\bm{V}$ is the right unitary matrix. Therefore, $\bm{\Phi}^{(n)}=\bm{A}_{(n)}^{\bm{\Phi}}\bm{A}_{(n)}^{\bm{\Phi}^\top} = \bm{U}^{(n)}\bm{\Sigma}\bm{V}^\top(\bm{U}^{(n)}\bm{\Sigma}\bm{V}^\top)^\top = \bm{U}^{(n)}\bm{\Sigma}\bm{V}^\top \bm{V}\bm{\Sigma}\bm{U}^{(n)\top} = \bm{U}^{(n)}\bm{\Sigma}^{2}\bm{U}^{(n)\top}$ since $\bm{V}^\top\bm{V}=\bm{I}$. It is known that $\bm{U}^{(n)^\top}\bm{U}^{(n)}=\bm{I}$, and $\bm{\Sigma}^{2}$ is a diagonal matrix whose diagonal entries are nonnegative. Thus, the left unitary matrix $\bm{U}^{(n)}$ from the singular value decomposition of $\bm{A}_{(n)}^{\bm{\Phi}}$ is equivalent to the eigenvector matrix of $\bm{\Phi}^{(n)}$.

\bibliographystyle{chicago}
\spacingset{1}
\bibliography{IISE-Trans}

\end{document}